\documentclass[journal,twoside,web]{ieeeconf}

\usepackage{cite}
\usepackage{amsmath,amssymb,amsfonts}
\usepackage{algorithmic}
\usepackage{graphicx}
\usepackage{textcomp}
\usepackage{multirow}
\usepackage{hyperref}
\usepackage[T1]{fontenc}
\usepackage{array, makecell}
\usepackage{threeparttable} % for table footnotes

\usepackage{geometry}
\def\BibTeX{{\rm B\kern-.05em{\sc i\kern-.025em b}\kern-.08em T\kern-.1667em\lower.7ex\hbox{E}\kern-.125emX}}

\begin{document}

\title{A Dataset and Benchmarks for Atrial Fibrillation Detection from Electrocardiograms of Intensive Care Unit Patients}

\author{Sarah~Nassar, Nooshin~Maghsoodi, Sophia~Mannina, Shamel~Addas, Stephanie~Sibley, Gabor~Fichtinger, David~Pichora, David~Maslove, Purang~Abolmaesumi, and~Parvin~Mousavi% <-this % stops a space
\thanks{© 2026 IEEE. Personal use of this material is permitted. Permission from IEEE must be obtained for all other uses. DOI: \href{https://ieeexplore.ieee.org/abstract/document/11614710}{10.1109/TBME.2026.3715145}}% <-this % stops a space
\thanks{S. Nassar is with the Department of Electrical and Computer Engineering at Queen's University, Kingston, ON, Canada.}% <-this % stops a space
\thanks{N. Maghsoodi and P. Mousavi are with the School of Computing at Queen's University, Kingston, ON, Canada.}% <-this % stops a space
\thanks{S. Mannina and S. Addas are with the Smith School of Business at Queen's University, Kingston, ON, Canada.}% <-this % stops a space
\thanks{S. Sibley is with the Departments of Emergency Medicine and Critical Care Medicine at Queen's University, Kingston, ON, Canada.}% <-this % stops a space
\thanks{G. Fichtinger is with the School of Computing and the Department of Electrical and Computer Engineering at Queen's University, Kingston, ON, Canada.}
\thanks{D. Pichora is with the Department of Surgery at Queen's University, Kingston, ON, Canada.}
\thanks{D. Maslove is with the Departments of Medicine and Critical Care Medicine at Queen's University, Kingston, ON, Canada.}% <-this % stops a space
\thanks{P. Abolmaesumi is with the Department of Electrical and Computer Engineering at the University of British Columbia, Vancouver, BC, Canada.}% <-this % stops a space
}

\maketitle

% ##################################################### %
% ##################################################### %
% ##################################################### %

\begin{abstract}
\textit{Objective:} Atrial fibrillation (AF) is the most common cardiac arrhythmia experienced by intensive care unit (ICU) patients and can cause adverse health effects. In this study, we publish a labelled ICU dataset and benchmarks for AF detection.
\textit{Methods:} We compared machine learning models across three data-driven artificial intelligence (AI) approaches: feature-based classifiers, deep learning (DL), and ECG foundation models (FMs). This comparison addresses a critical gap in the literature and aims to pinpoint which AI approach is best for high-performing AF detection. Electrocardiograms (ECGs) from a Canadian ICU and the 2021 PhysioNet/Computing in Cardiology Challenge were used to conduct the experiments. Multiple training configurations were tested, ranging from zero-shot inference to transfer learning. \textit{Results:} Across both datasets, ECG FMs generally performed best, followed by DL, then feature-based classifiers. However, the difference between DL and feature-based classifiers is minimal and highly dependent on the model selected, with feature-based classifiers obtaining a very slightly higher average performance on our ICU test set but DL achieving a higher overall maximum performance. The models that achieved the top F1 score on our ICU test set were InceptionV3 with recurrence plots and a fine-tuned ECGFounder (F1=0.88). \textit{Conclusion:} This study demonstrates promising potential for using AI to build an automatic patient monitoring system. \textit{Significance:} By publishing our labelled ICU dataset and performance benchmarks, this work enables the research community to continue advancing the state-of-the-art in AF detection in the ICU environment.~\url{https://physionet.org/content/kingston-icu-af/}
\end{abstract}


% ##################################################### %
% ##################################################### %
% ##################################################### %

\section{Introduction}

\subsection{Motivation}

Atrial fibrillation (AF) is the most common cardiac arrhythmia and can lead to negative health outcomes such as heart failure and stroke~\cite{kornej2020epidemiology}. The prevalence of AF is higher in the intensive care unit (ICU) than in the general population at up to 15-20\% or more~\cite{moss2017new, paula2024atrial, rottmann2024atrial}. While ICU patients may present with pre-existing AF, they are also at an increased risk of developing new-onset AF due to critical illness~\cite{bosch2018atrial}. AF is primarily diagnosed by visually inspecting the electrocardiogram (ECG) reading of a patient and identifying morphological irregularities such as inconsistent intervals between R peaks and missing P waves~\cite{hindricks20212020}. AF management in ICU patients presents a unique challenge as these patients are at higher risk of rapid health deterioration. However, ICU patients are connected to bedside monitors that continuously capture their ECG readings, allowing for automatic monitoring of their cardiac rhythm. Therefore, it is of utmost importance to leverage these continuous telemetry data to allow for timely initiation of treatment strategies.

Given the continuous nature of ECG data capture, the amount of time, training, and experience needed to interpret ECGs, and the urgency required to treat patients with AF as soon as possible, artificial intelligence (AI) can be used to develop an automatic detection algorithm that can continuously process a patient's ECG and classify whether AF is occurring. To facilitate this downstream application of real-time patient monitoring, different AI technologies need to be compared to find the best approach suitable for the ICU context.

% ##################################################### %

\subsection{Problem}

The body of literature in AI-powered ECG-based arrhythmia detection is rich, with many diverse approaches. By exploring existing literature, it can be noted that deep learning (DL) is currently one of the most common approaches~\cite{boulif2023literature}. Classical machine learning (ML) approaches, which require hand-crafted features to be extracted from ECGs, have recently become less common, even though the ECG signal modality has distinguishable features that clinicians use to identify arrhythmia and that can be quantified. Additionally, classical ML is typically less resource-intensive than DL, which is an important consideration for real-time deployment contexts. Further, ECG foundation models (FMs) are an emerging trend~\cite{mckeen2025ecg, li2025electrocardiogram}. FMs are large deep neural networks that have been pretrained on large volumes of data and can be adapted to learn related tasks with minimal additional training and data. By observing the abundance of DL research for arrhythmia detection, one may instinctively presume it would perform best at AF detection in the ICU, but this has been rarely systematically tested, nor have different approaches to DL and newly published ECG FMs been included in such a comparison. Therefore, a limitation in the current literature is the absence of direct and comprehensive comparisons of different AF detection approaches for data from ICU patients.

Aside from research in arrhythmia detection more generally, the literature focusing on AF detection in the ICU is minimal. Therefore, another limitation in the current literature is the scarcity of AF detection studies in the ICU context, especially with a dataset spanning a large enough number of patients. The ICU is a unique environment with noisy signals due to patient movement and prevalent alarms that can lead to alarm fatigue in care providers, making the implementation of high-performing modelling approaches with minimal false alarms essential.

The latest research frontier is the future prediction of AF, which aims to forecast the occurrence of future episodes of AF as opposed to flagging AF only if it is currently occurring. Being able to anticipate an AF episode ahead of time could allow critical care clinicians to initiate proactive and preventive measures. To enable the training of AI models for AF forecasting, a labelled dataset would need to be curated by identifying AF onsets from long-term ECG recordings. This data labelling is not a trivial task since it is not feasible for an expert annotator to process hours or days of data. Therefore, having high-performing AF detection models can support pinpointing AF onsets by obtaining model classifications across entire patient stays and using them as non-expert weak labels for AF forecasting.

% ##################################################### %

\subsection{Objectives and Contributions}

The goal of this research is to address gaps in the current literature in AF detection, in particular in the ICU context, and enable future research in AF forecasting. The main contributions of this study are as follows:

\begin{enumerate}
    \item We performed a comprehensive benchmark of AF detection performance with a wide variety of machine learning models within three data-driven AI approaches, encompassing feature-based classifiers, deep learning, and foundation models, and across multiple training configurations. This comparison involves the large publicly available 2021 PhysioNet dataset and our institutional ICU dataset.
    \item We published our institutional ICU dataset with almost 600 labelled 10~s four-lead ECGs from unique patients from Kingston, Ontario, Canada to facilitate further research by the wider research community.
    \item As a result of comparing several models across different AI approaches and training configurations, we demonstrated that fine-tuning an ECG foundation model improved the AF detection performance on our ICU dataset compared to previous research~\cite{chen2021detecting, chen2022quantifying, chen2023deep}.
\end{enumerate}

% ##################################################### %
% ##################################################### %
% ##################################################### %

\section{Related Work}

\subsection{Arrhythmia Classification Approaches and Research Trends}

In the context of ECG-based arrhythmia detection, there are two main approaches under the umbrella of AI to address the task: classical ML and DL. Classical ML algorithms are typically used with a limited set of pre-defined features that are extracted from ECG signals, including time domain features (e.g., time between peaks), spatial features (e.g., peak amplitude), non-linear features (e.g., entropy), or frequency or time-frequency features~\cite{jahan2022short}. On the other hand, DL approaches typically use one-dimensional (1-D) convolutional architectures applied to the raw single- or multi-lead ECG signals~\cite{hannun2019cardiologist, ribeiro2020automatic}~or two-dimensional (2-D) architectures applied to images~\cite{izci2019cardiac}~or signal-to-image transformations (e.g., spectrograms)~\cite{lekkas2025time}~of the ECG recordings, such that both the feature extraction and arrhythmia classification steps are end-to-end automated during training. More recently, a third approach has emerged: ECG FMs~\cite{mckeen2025ecg, li2025electrocardiogram}, which are deep neural networks that have been pretrained with large volumes of raw ECG signals and can be used in either a zero-shot fashion with no additional training or by fine-tuning them to a downstream task with minimal additional data.

There exist several review papers covering research trends in the use of AI for arrhythmia detection. Boulif et al. found 40 relevant articles for ECG-based arrhythmia diagnosis that were published from January 2010 to September 2022~\cite{boulif2023literature}. The majority of these articles (up to 75\%) applied DL and the rest either used classical ML or combined both methods. The most common DL architecture was that of the convolutional neural network (CNN) or a hybrid method involving a CNN. The 2021 PhysioNet/Computing in Cardiology (CinC) Challenge, which asked participants to explore the number of ECG leads needed to accurately classify cardiac abnormalities, received 39 official entries~\cite{reyna2022issues}. Of these, 64\% approached the challenge with DL, including CNNs in general and ResNet-based approaches in particular, 10\% approached it with feature-based ML classifiers, and 26\% approached the challenge by combining hand-crafted extracted features with DL. So far, the publication in AI-based arrhythmia classification with the most citations describes a 1-D CNN architecture published by researchers in the Stanford Machine Learning Group~\cite{hannun2019cardiologist}. As observed from the review paper, the challenge entries, the most-cited paper, and the fact that many other review papers only focus on DL, it can be said that most of the research in this area approaches the task with DL.

Unlike the literature review with a wider scope by Boulif et al., Ansari et al. found 78 papers from January 2017 to January 2023 that specifically used DL for ECG-based arrhythmia detection and/or classification and that reported an accuracy of 96\% or higher~\cite{ansari2023deep}. While being a more recent review, it does not aggregate statistics about the different DL approaches and datasets used. Xiao et al. searched for papers involving ECG-based arrhythmia classification with DL and found 368 studies that were published until December 2022, 59\% of which used a CNN architecture and at least 13\% of which used a hybrid model involving the CNN architecture~\cite{xiao2023deep}. Additionally, 61\% of the papers used the MIT-BIH Arrhythmia Database dataset, which only covers 47 subjects.

Recently, there have been several FMs pretrained for ECG analysis with public weights and zero-shot capability, including ECG-FM and ECGFounder. ECG-FM is a Transformer-based model pretrained using a hybrid self-supervised learning approach with 1.4 million 10~s 12-lead ECGs from the Medical Information Mart for Intensive Care IV ECG (MIMIC-IV-ECG) and 2021 PhysioNet Challenge datasets~\cite{mckeen2025ecg}. ECGFounder is a CNN-based model pretrained with over 7.5 million 10~s 12-lead ECGs from the Harvard-Emory ECG Database~\cite{li2025electrocardiogram}.

There are many more studies addressing arrhythmia detection with DL than with classical ML, reflecting the general trend in the AI research field. However, to our knowledge, there is yet to be a clear determination of which approach performs best through a comprehensive comparison, especially for AF detection in the ICU. Therefore, the application of classical ML methods was not dismissed in this research.

% ##################################################### %

\subsection{AF Detection Research Trends} \label{section:related_work_2}

In contrast to arrhythmia detection, there are few reviews that specifically focus on AF detection. Murat et al. looked for DL-based AF detection studies, but not focused on the critical care context, and found 24 relevant papers published from 2017 to September 2021~\cite{murat2021review}. Once again, the CNN architecture was the most frequently appearing architecture (either alone or with another architecture) in 58\% of the papers. This review focused on DL approaches, and did not include other AI approaches in its scope.

Although there is an abundance of research in AI-driven arrhythmia detection, the amount of work addressing AF detection specifically in critical care contexts is lacking. We were only able to find one recent systematic review looking for research related to new-onset AF detection in the ICU from late 2020 to early 2023 which claimed to have found no more than three articles, two of which used classical ML and one of which used DL, specifically a CNN model~\cite{glaser2024machine}. The study using a CNN model was one of the previous works published by our research team involving our institutional ICU data~\cite{chen2022quantifying}. In contrast to a standard 12-lead ECG that typically captures a 10~s snapshot of the heart's electrical activity while the subject remains still, telemetry monitoring in the ICU captures a reduced subset of three to five leads over the bedridden patient's entire stay~\cite{francis2016ecg}. Ambulatory ECGs similarly capture a small subset of leads over a prolonged period of time, but in an outpatient setting while the subject performs their normal daily activities. These differences between ICU ECGs, standard 12-lead ECGs, and ambulatory ECGs warrant a focused study into AF detection specifically for the unique ICU setting.

There have been other studies that compared the performance of various AI methods for AF detection. For example, a study compared the performance of using ML algorithms with manual feature extraction and using DL methods~\cite{liaqat2020detection}. With the 2017 PhysioNet Challenge dataset, all DL models outperformed the ML models for the AF class. However, the authors do not clarify whether for the DL models, they used raw signals or images as input. As another example, a submission to the 2017 PhysioNet Challenge compared a feature-based ML classifier and a DL CNN for AF detection~\cite{andreotti2017comparing}. With 5-fold cross-validation (CV) on the training set, the feature-based ML models outperformed their deep learning CNN counterparts for the AF class. However, this trend differed on the challenge test set and the two approaches generally performed similarly. In this study, our aim is to leverage the newer and larger 2021 PhysioNet Challenge dataset, incorporate a bigger set of AI models, clearly distinguish between the signal-based and image-based DL models, test signal-to-image transformation methods, include newly published FMs in the comparison, and focus on the ICU environment.

% ##################################################### %
% ##################################################### %
% ##################################################### %

\section{Methods}

Fig.~\ref{figure:overview}~illustrates the framework designed for the comparison. Two datasets were used: a small ICU-specific dataset and a large non-ICU-specific dataset. Several models within three AI approaches were trained: feature-based models, signal-based and image-based DL models, and ECG FMs. These models were compared across four training configurations with varying sample sizes. For feature-based models, we extracted many signal-derived features (Table~\ref{table:features}) and tested almost all classical ML algorithms. For DL, a limited number of models were selected due to the longer training time required. These models aligned with major trends in the literature for input formats, such as raw signals, images, and signal-to-image transformations (e.g., spectrograms, scaleograms, and recurrence plots). For FMs, all publicly available models that we could download and fine-tune at the time of writing this paper were tested.

\begin{figure*}
\centering
\includegraphics[width=0.75\textwidth]{figures/overview.png}
\caption{Overview of comparison framework showing the two datasets, three AI approaches, and four training configurations. Each training configuration's card showcases icons corresponding to the applicable AI approach(es), the training set used with increasing sample size from left to right, and the test set(s) used for performance evaluation.}
\label{figure:overview}
\end{figure*}

\subsection{Datasets, Preprocessing, and Feature Extraction}

Two datasets were used in this research: our institutional ICU dataset and the publicly available 2021 PhysioNet/CinC Challenge dataset~\cite{reyna2021will, goldberger2000physiobank}. Our institutional ICU dataset was collected from a tertiary hospital in Kingston, Ontario, Canada. It includes archival data from 2015 to 2020 from bedside monitors covering 1,043 de-identified patients. The ECG telemetry leads (I, II, III, and a V1 equivalent) were sampled at 240~Hz. The collection and usage of these data were granted ethical clearance by the Queen's University Health Sciences and Affiliated Teaching Hospitals Research Ethics Board (File Number: 6024689). The need for informed consent was waived because the data were already being collected as part of routine clinical practice and stored in de-identified format. The data collection and labelling procedures are described in more detail in~\cite{chen2021deep}. In summary, we used ECGs from the first round of labelling, whereby, from a subset of patients, a randomly selected 10~s ECG was annotated by two critical care physicians and a third physician was consulted in the case of ties. This resulted in 613 relevant ECGs from unique patients where 513 were labelled as sinus rhythm (regardless of the underlying heart rate) and 100 were labelled as AF (combined with atrial flutter). ECGs with other labels (pacemaker, bigeminy and trigeminy, ventricular tachycardia and fibrillation, other tachycardia, and other bradycardia), which had low or no prevalence, are not considered here.

The 2021 PhysioNet Challenge dataset is a multi-site collection of labelled 12-lead ECG data. Our study uses the training set from the challenge, which is the only set available for download from PhysioNet. This training set mainly contains data from four sources: the Chapman University, Shaoxing People's Hospital, and Ningbo First Hospital database (hereafter referred to as Chapman/Shaoxing \& Ningbo), the 2018 China Physiological Signal Challenge (CPSC) dataset and unused data from the challenge (hereafter referred to as CPSC\_2018 \& CPSC\_2018\_Extra), the Georgia 12-lead ECG Challenge database (hereafter referred to as Georgia), and the Physikalisch-Technische Bundesanstalt (PTB) and PTB-XL databases (hereafter referred to as PTB \& PTB-XL).

For this multi-site public dataset, ECG recordings that are shorter than 10~s were excluded and those longer were trimmed to only include the first 10~s of data. Each ECG recording has one or more corresponding label(s) according to the Systematized Nomenclature of Medicine Clinical Terms (SNOMED CT) system. The SNOMED CT codes for sinus rhythm, sinus bradycardia, sinus tachycardia, and sinus arrhythmia were mapped to the sinus rhythm class, and the codes for AF, atrial flutter, AF and atrial flutter, chronic AF, and rapid AF were mapped to the AF class. ECG recordings with both class labels or with one or more co-occurring labels from the unused labels in our institutional data mentioned above (pacemaker, bigeminy and trigeminy, ventricular tachycardia and fibrillation, other tachycardia, and other bradycardia) were removed. ECG recordings with one or more flat or missing leads were removed. The remaining ECG signals were bi-directionally filtered with a second order Butterworth bandpass (5-30~Hz) filter, and the signals from the public dataset were downsampled to 240~Hz to match the sampling frequency of our ICU dataset.

For feature extraction, the first step was to detect the locations of the R peaks with the SleepECG Python package~\cite{SleepECG}, which uses a modified version of the beat detection algorithm with adaptive thresholds described in~\cite{pan1985real}. From these R peak locations, the NeuroKit2 package~\cite{makowski2021neurokit2}~was used to estimate the signal quality and calculate the heart rate along with various heart rate variability (HRV) features and a feature to quantify P peak missingness. The full set of 27 extracted features per lead is listed in Table~\ref{table:features}.

\begin{table}
\centering
\caption{Set of features extracted per ECG lead.}
\label{table:features}
\resizebox{\columnwidth}{!}{
\begin{tabular}{|l|l|}
\hline
\multirow{2}{*}{General features} & Signal quality \\ \cline{2-2} & Heart rate in beats per minute \\ \hline
\multirow{8}{*}{HRV time features} & Minimum, maximum, mean, and median RR interval \\ \cline{2-2} & \makecell[l]{Standard deviation, median absolute deviation, interquartile range, \\ and 20th and 80th percentiles of RR intervals} \\ \cline{2-2} & \makecell[l]{Standard deviation of successive differences between RR intervals} \\ \cline{2-2} & \makecell[l]{Percentage of absolute differences in successive RR intervals \\ greater than 20 ms and 50 ms} \\ \cline{2-2} & \makecell[l]{HRV triangular index measuring number of RR intervals divided \\ by height of RR interval histogram} \\ \hline
\multirow{6}{*}{HRV non-linear features} & \makecell[l]{\textbf{Poincaré plot basic features:} Standard deviation perpendicular to \\ and along line of identity and associated area of ellipse} \\ \cline{2-2} & \makecell[l]{\textbf{Poincaré plot asymmetry features:} Short-term, long-term, and \\ total variance of contributions of decelerations and accelerations} \\ \cline{2-2} & \makecell[l]{\textbf{Heart rate fragmentation feature:} Percentage of inflection \\ points in RR intervals} \\ \cline{2-2} & \textbf{Entropy feature:} Shannon entropy \\ \hline
\multirow{1}{*}{P peak missingness feature} & Average P peak amplitude \\ \hline
\end{tabular}}
\end{table}

ECG segments with any missing lead (defined as three or less R peaks detected) were dropped due to the inability to calculate other features. The features were concatenated such that each 10~s ECG recording has a single row containing four columns for each feature extracted from each of the four leads. This results in 108 columns for each ECG recording. Although features across leads are highly correlated, each lead provides valuable information and was retained given the ICU's susceptibility to signal noise and imperfect lead placement. By the end of the data processing stage, the remaining record breakdown and AF class prevalence are shown in Table~\ref{table:dataset}.

The MIMIC-IV-ECG database~\cite{gow2023mimic}, which contains about 800,000 ECGs from nearly 160,000 ICU patients with more than 600,000 cardiologist reports, has yet to publish the de-identified free-text cardiologist reports. Therefore, to our knowledge, there is no other open-source dataset with labelled ECGs from ICU patients.

\begin{table}
\centering
\caption{Breakdown of the two datasets.}
\label{table:dataset}
\resizebox{\columnwidth}{!}{
\begin{tabular}{|c|c|c|c|}
\hline
\multicolumn{1}{|c|}{Dataset} & Portion & \# of ECGs & \makecell{AF Class \\ Prevalence} \\ \hline
\multirow{1}{*}{Kingston ICU} & Labelled & 596 & 17\% \\ \hline
\multirow{5}{*}{\makecell{2021 PhysioNet \\ Challenge~\cite{reyna2021will, goldberger2000physiobank}}} & Overall & 70,363 & 18\% \\ \cline{2-4} & Chapman/Shaoxing \& Ningbo & 41,331 & 22\% \\ \cline{2-4} & CPSC\_2018 \& CPSC\_2018\_Extra & 2,687 & 53\% \\ \cline{2-4} & Georgia & 5,623 & 12\% \\ \cline{2-4} & PTB \& PTB-XL & 20,722 & 7\% \\ \hline
\end{tabular}}
\end{table}

% ##################################################### %

\subsection{AI Approaches}

\subsubsection{Feature-Based Models}

The first AI approach used for AF detection is classical ML with the features that were extracted from the ECG recordings and formatted in a tabular layout. The models include k-nearest neighbours (KNN), multi-layer perceptron (MLP), support vector machine (SVM), logistic regression (LR), and four tree-based ensembles with 1,000 estimators each: random forest (RF), Light Gradient-Boosting Machine (LightGBM)~\cite{ke2017lightgbm}, Extreme Gradient Boosting (XGBoost)~\cite{chen2016xgboost}, and Categorical Boosting (CatBoost)~\cite{prokhorenkova2018catboost}. Prior to the model training stage, features were min-max scaled. In addition, the zero-shot performance of TabPFN v2 (Tabular Prior-data Fitted Network), a transformer-based foundation model for tabular data, was tested~\cite{hollmann2025accurate}. TabPFN v2 automatically handles preprocessing, which involves robust scaling. Although the model is pretrained and available for use in zero-shot fashion, it needs a training dataset to work in an in-context learning paradigm. The purpose of this training dataset is not to fine-tune the model's weights, but to allow the model to use it as context for inference with new data.

% ##################################################### %

\subsubsection{DL Models}

The second AI approach is DL with 1-D and 2-D CNNs. \underline{1-D CNNs with Raw ECG Signals:} The first DL technique is to use 1-D CNNs that accept the raw ECG signals. While these models are referred to as 1-D CNNs since they process temporal sequences through 1-D convolutions, the signals provided as input are four-lead ECGs, with leads presented to the models as four channels. Two CNN architectures were used: the first (Goodfellow CNN) is similar to the one that was used in previous research involving our institutional ICU data and is based on~\cite{goodfellow2018towards}~while the second (Stanford CNN) is an architecture designed by the Stanford Machine Learning Group and is, so far, the most-cited publication in AI-based arrhythmia classification~\cite{hannun2019cardiologist}. In our group's previous research, one way the Goodfellow CNN was trained was using 2.5~s ECG segments extracted from the full 10~s recordings. For inference with this method, a sliding window (with overlap of half the window length) of model output probabilities are averaged to come up with a final classification. In this work, this method was compared with the same model trained on the full 10~s recordings. For the Stanford CNN, the ECG recordings were limited to 9.6~s and the window size used was about 1.1~s because the architecture is designed to process inputs with a length of 256 or a multiple of 256. ECG signals were min-max scaled lead-wise prior to model training. \underline{2-D CNNs with ECG Images:} The second DL technique is to use 2-D CNNs that take images of the ECGs as input. Two well-known architectures were trained: ResNet-18~\cite{he2016deep}~and Inception v3~\cite{szegedy2016rethinking}. The ECG plot package~\cite{ecgplot}~was used to plot ECG recordings, which were min-max scaled lead-wise prior to plotting. The image size is 319$\times$999. Each channel was min-max scaled prior to being passed to the model for processing. \underline{2-D CNNs with Signal-to-Image Transformations:} The third DL technique is to use 2-D CNNs that take images obtained through signal-to-image transformations as input. Three types of images were used: spectrograms generated with the short-time Fourier transform, scaleograms generated with the continuous wavelet transform, and recurrence plots. Since there are four applicable ECG leads, the signal-to-image transformations resulted in four images per ECG recording. These images, in grayscale format, were combined as channels for the 2-D CNN. To do this, the first convolutional layer of the models was changed to accept four channels instead of the usual three. The image size is 319$\times$999. Each channel was min-max scaled prior to being passed to the model for processing.

% ##################################################### %

\subsubsection{FMs}

The third AI approach is to use open-weight ECG FMs that accept the raw ECG signals as input. Two such models were used: the transformer-based ECG-FM~\cite{mckeen2025ecg}~and the CNN-based ECGFounder~\cite{li2025electrocardiogram}. With pretrained FMs, the data preprocessing for downstream tasks should typically follow the same procedure as the pretraining stage. ECG-FM accepts 5~s signals with 12 leads of data sampled at 500~Hz and no signal filtering. A windowing procedure was again performed to aggregate model outputs across the full 10~s. ECGFounder accepts 10~s signals with either 12 leads or a single lead of data sampled at 500~Hz and a three-step filter composed of a high-pass filter with a cutoff frequency of 1~Hz, a second-order 30~Hz Butterworth low-pass filter, and a bandstop (50-60~Hz) filter.

% ##################################################### %

\subsection{Experimental Setup for Comparison Framework}

The Kingston ICU dataset was split into two halves to yield a training set (n=298) and a test set (n=298). To address class imbalance, this split was done in a stratified fashion to ensure the same ratio of AF samples remained in both sets. For the 1-D and 2-D CNN training, the learning rate was set to 0.05. For the ECG FM fine-tuning, the learning rate was reduced to 0.0001. The batch size was either 32, 64, or 128 depending on the training data size and model memory requirements, and the models were optimized with binary cross-entropy loss and the Adam optimizer~\cite{kingma2014adam}. The ICU training set was used for early stopping when the loss failed to improve for several consecutive epochs, up to a maximum of 200 epochs. Finally, the classification threshold was set of 0.5.

% ##################################################### %

\subsubsection{Training Config \#1: Zero-Shot Inference (n=0)}

The first training configuration is to use zero-shot ECG FMs for inference with no additional training. Two fine-tuned versions of ECG-FM that were published were tested. The fine-tuning data for the first version (ECG-FM v1) came from the 2021 PhysioNet Challenge and the fine-tuning data for the second version (ECG-FM v2) came from the MIMIC-IV-ECG dataset. The preprocessing required to test ECG-FM involved resampling using linear interpolation (if needed) to match the required sampling frequency of 500~Hz, standard scaling (i.e., Z-score normalization), setting unmeasured leads to zero, and no filtering. Two zero-shot versions of ECGFounder that were published were also tested: the 12-lead ECGFounder with unmeasured leads set to zero and the single-lead ECGFounder with the predictions across all available leads averaged. The preprocessing required to test ECGFounder involved resampling with linear interpolation (if needed) to match the required sampling frequency of 500~Hz, filtering, and standard scaling. The fine-tuned ECG-FM and ECGFounder models output multi-label classifications for either 17, 26, or 150 classes. To convert the multi-label classification output format into a binary classification format, we applied the sigmoid transformation on the outputs for AF and atrial flutter and used the maximum of these two as the positive class probability to perform thresholding and calculate performance metrics.

% ##################################################### %

\subsubsection{Training Config \#2: Train on ICU Data (n=298)}

The second training configuration is to train all models with the ICU training set. Although small in size, it is an ICU-specific dataset representing the main context of interest. Most models were trained from scratch. However, the pretrained ECG FMs were fine-tuned. For both FMs, the input layer was modified to accept four leads instead of 12, while keeping the pretrained weights corresponding to the four relevant leads.

% ##################################################### %

\subsubsection{Training Config \#3: Train on PhysioNet Data (n=70,363)}

The third training configuration is to train all models with the 2021 PhysioNet Challenge data. Due to the small size of the ICU data, the utility of using a large but non-ICU-specific dataset for training was investigated. To evaluate these models on the PhysioNet data and avoid patient data leakage within sites, leave-site-out CV was performed. Once again, most models were trained from scratch while the pretrained ECG FMs were fine-tuned.

% ##################################################### %

\subsubsection{Training Config \#4: Transfer Learning (n=70,363+298)}

The fourth training configuration is to perform transfer learning, which can help tackle the problem of not having a sufficient number of annotated in-house training samples for DL. This concept has demonstrated improved performance in a research study for ECG classification compared to training a model from scratch with a small dataset~\cite{weimann2021transfer}. For each DL model trained on the full 2021 PhysioNet Challenge data and each ECG FM fine-tuned on these data, the network was further fine-tuned with the ICU training set.

% ##################################################### %
% ##################################################### %
% ##################################################### %

\section{Results and Discussion}

This section documents the classification metrics for each model across all training configurations on the ICU test set. Only a summary of the performance on the 2021 PhysioNet Challenge data is shown here. Although comparing AI approaches on the non-ICU PhysioNet dataset is not the main focus of this study, we present a summary of results as a form of external validation to show consistency of trends on more than one dataset given that the lack of comprehensive comparative studies for AI-driven AF detection in the literature spans both the ICU and non-ICU domains.

The calculated metrics include recall/sensitivity, precision/positive predictive value, and F1 score. Recall focuses on a model's ability to correctly flag the positive cases, while precision focuses on a model's ability to minimize making false positive predictions. To detect AF in ICU patients, recall is an important metric to maximize, but at the same time, a model needs to have a high precision in order to avoid exacerbating alarm fatigue in care providers, which can damage trust in the model and hinder the effectiveness of its use. To balance between the need to have a high recall as well as a high precision, the main metric used to compare the performance of the different models trained in this work is the F1 score, which is the harmonic mean of both recall and precision. A larger set of evaluation metrics is reported in Tables~S1~to~S4~in the Supplementary Materials.

\subsection{Training Config \#1: Zero-Shot Inference (n=0)}

\begin{figure}
\centering
\includegraphics[width=\columnwidth]{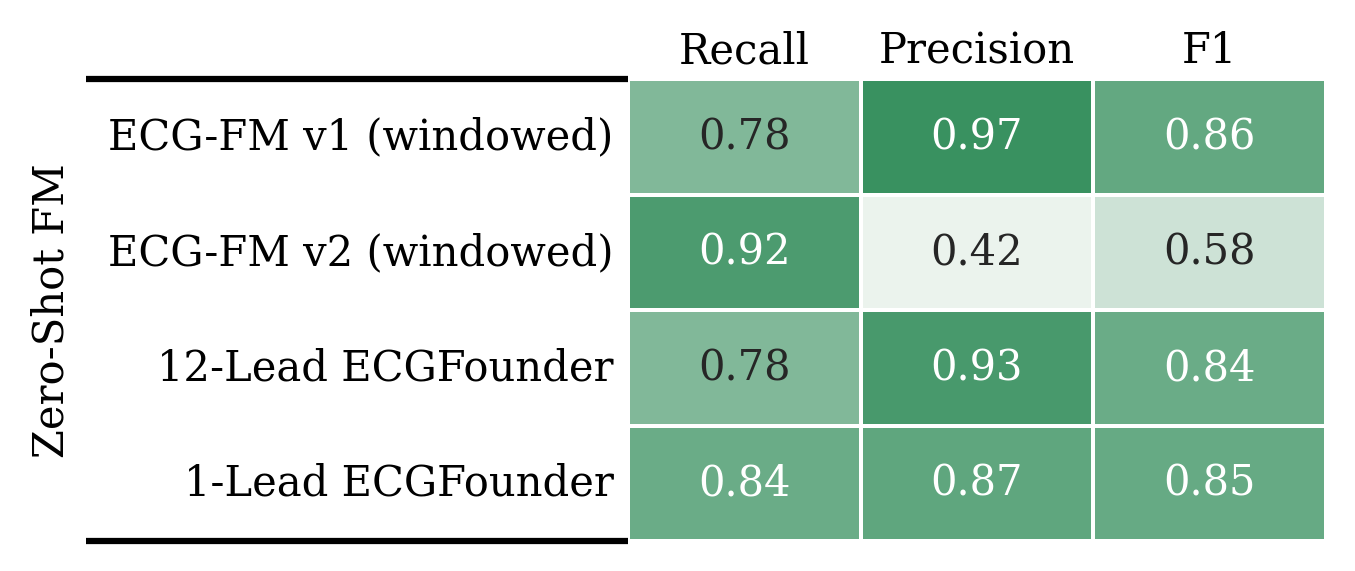}
\caption{ICU test set performance of zero-shot ECG FMs (training config \#1). Darker table cells indicate better performance.}
\label{table:tc1_icutest}
\end{figure}

Fig.~\ref{table:tc1_icutest}~shows the zero-shot performance of ECG-FM v1, ECG-FM v2, the 12-lead ECGFounder, and the single-lead ECGFounder (with predictions averaged across the four available leads) on the ICU test set. The best F1 score of 0.86 was achieved by ECG-FM v1. This F1 score is a good starting point for a zero-shot model with no additional training. ECG-FM v1 outperformed ECG-FM v2 across all metrics except recall. A potential explanation could be that the MIMIC-IV-ECG dataset used to fine-tuned ECG-FM v2 contains labels parsed from free-text ECG reports that were automatically generated by machine measurements as opposed to expert physicians. Depending on the accuracy of these machine measurements and the text parsing algorithm, it is possible that ECG-FM v2 was fine-tuned with inaccurate labels. Both the 12-lead ECGFounder and single-lead ECGFounder have a very similar F1 score. The single-lead ECGFounder may be slightly outperforming the 12-lead version because our ICU dataset only contains four leads and we set unmeasured leads to zero in order to test the 12-lead model. We tested other approaches to fill the missing lead information (e.g., by setting them all to lead I), but this either worsened or did not significantly change performance.

% ##################################################### %

\subsection{Training Config \#2: Train on ICU Data (n=298)}

\begin{figure}
\centering
\includegraphics[width=\columnwidth]{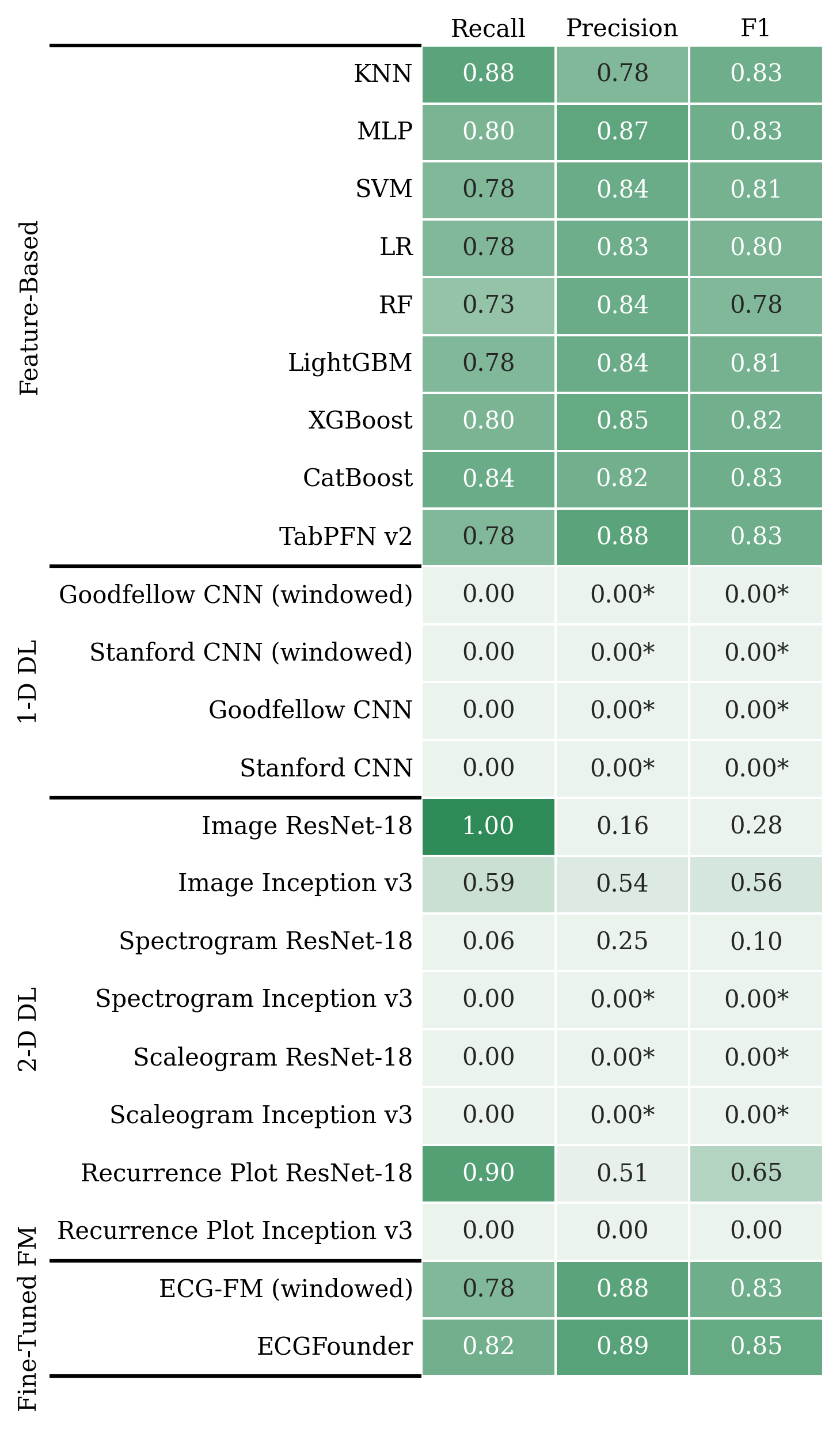}
\caption{ICU test set performance of models trained on the ICU training set (training config \#2). *Precision and F1 score are mathematically undefined due to a zero denominator (i.e., zero positive predictions), but are conventionally bound to 0.00.}
\label{table:tc2_icutest}
\end{figure}

Fig.~\ref{table:tc2_icutest}~shows the performance metrics of the models trained with the ICU training set and evaluated on the ICU test set. The best F1 score of 0.85 and best precision of 0.89 were achieved by a fine-tuned ECGFounder. All the feature-based models and both fine-tuned ECG FMs have a higher F1 score and precision than all the 1-D and 2-D DL models trained from scratch. As expected, DL models with a large number of parameters that are trained from scratch fail to learn meaningful features with a very small amount of training data, and frequently make all-negative or all-positive predictions resulting in zero or undefined metrics. On the other hand, the feature-based models and fine-tuned ECG FMs which were previously pretrained are robust to the small training data size.

% ##################################################### %

\subsection{Training Config \#3: Train on PhysioNet Data (n=70,363)}

\begin{figure}
\centering
\includegraphics[width=\columnwidth]{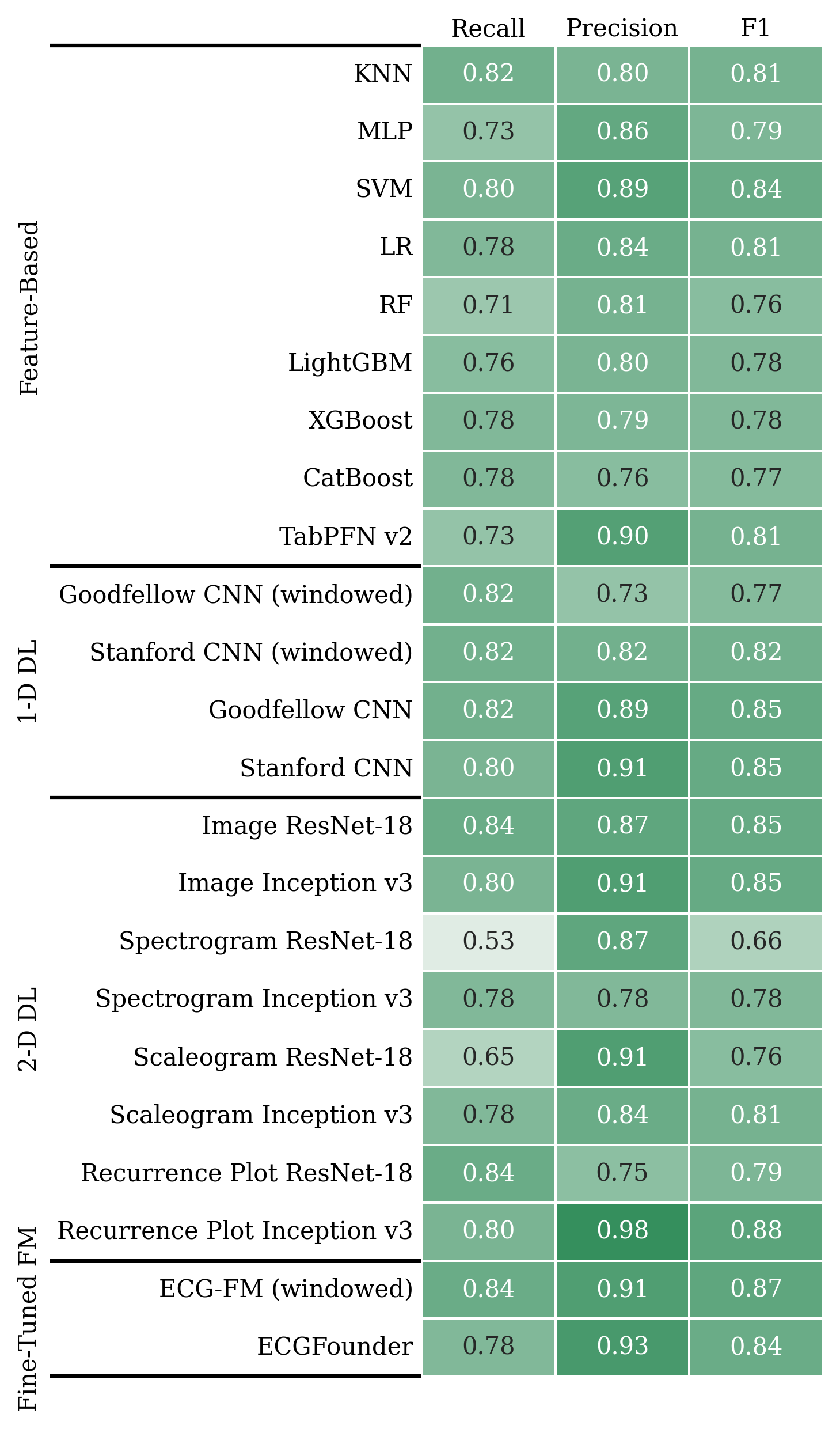}
\caption{ICU test set performance of models trained on the 2021 PhysioNet Challenge data (training config \#3).}
\label{table:tc3_icutest}
\end{figure}

Fig.~\ref{table:tc3_icutest}~shows the performance metrics of the models trained on the 2021 PhysioNet Challenge data and evaluated on the ICU test set. The best F1 score of 0.88 and best precision of 0.98 were achieved by Inception v3 with recurrence plots. The best recall of 0.84 was achieved by ResNet-18 with ECG images or recurrence plots and by a fine-tuned ECG-FM. Because the PhysioNet dataset contains tens of thousands of samples, many of the 1-D and 2-D DL models were able to surpass the feature-based models. The fine-tuned ECG FMs also out-performed the feature-based models. For the 1-D CNNs, there is a clear improvement in using larger input sizes to allow the models to process longer signals. The Goodfellow CNN's F1 score increases from 0.77 to 0.85 by increasing the input size from 2.5~s to 10~s, and the precision jumps from 0.73 to 0.89. The Stanford CNN's F1 score increases from 0.82 to 0.85 by increasing the input size from 1.1~s to 9.6~s, and the precision jumps from 0.82 to 0.91. In general, the signal-to-image transformation 2-D CNNs do not have a very high performance with the exception of Inception v3 with recurrence plots.

% ##################################################### %

\subsection{Training Config \#4: Transfer Learning (n=70,363+298)}

\begin{figure}
\centering
\includegraphics[width=\columnwidth]{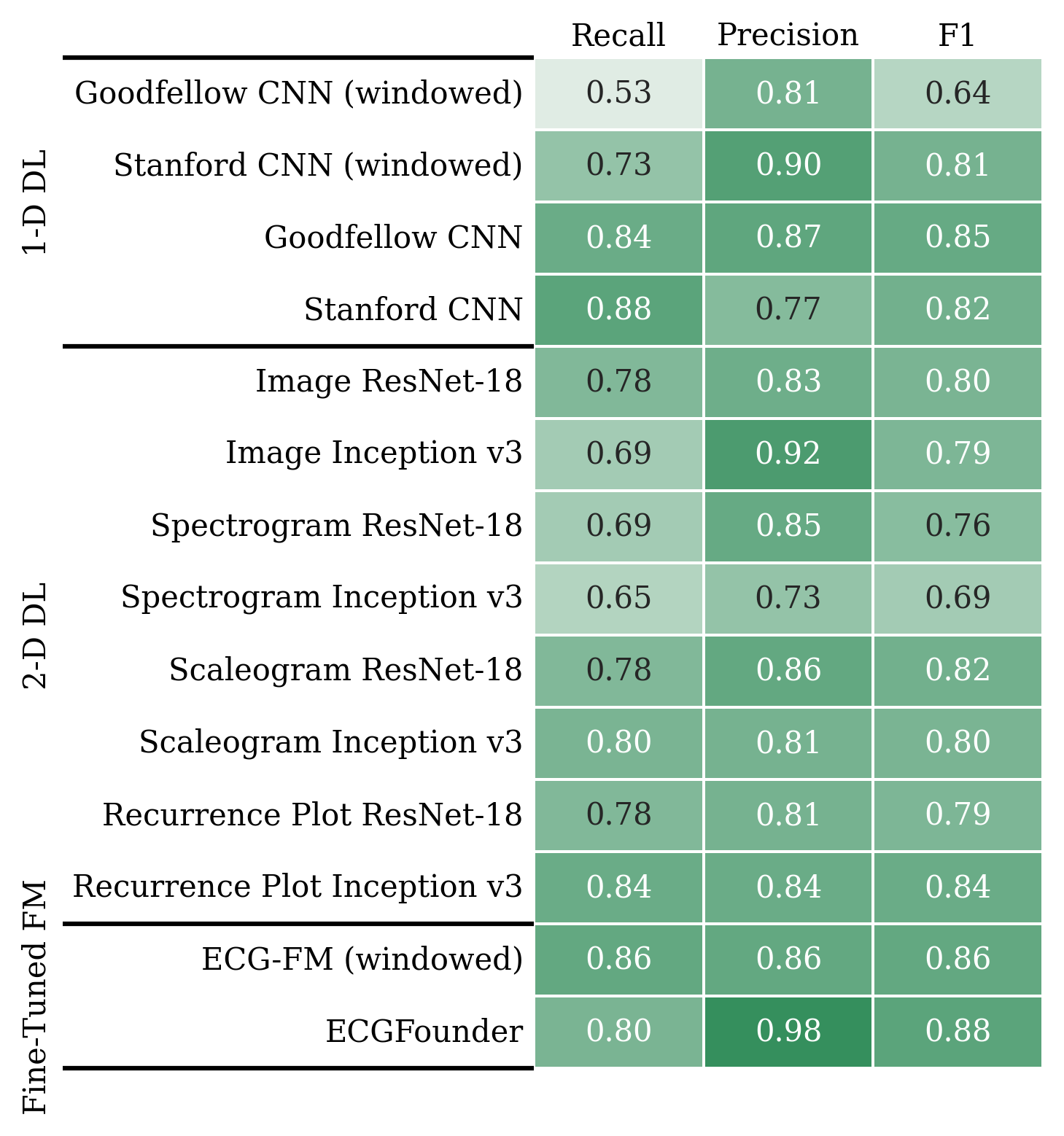}
\caption{ICU test set performance of models trained on the 2021 PhysioNet Challenge data then fine-tuned with the ICU training set for transfer learning (training config \#4).}
\label{table:tc4_icutest}
\end{figure}

Fig.~\ref{table:tc4_icutest}~shows the performance metrics of the DL models and ECG FMs first trained or fine-tuned on the 2021 PhysioNet Challenge data, then fine-tuned for transfer learning with the ICU training set, and finally evaluated on the ICU test set. The best F1 score of 0.88 and best precision of 0.98 was achieved by ECGFounder. The best recall of 0.88 was achieved by the 1-D Stanford CNN with 9.6~s inputs. Transfer learning did not prove to be advantageous compared to the previous training configuration, except when it comes to ResNet-18 with spectrograms or scaleograms and ECGFounder.

A figure summarizing the progression of the F1 score across all four training configurations on the ICU test set can be found in the Supplementary Materials.

% ##################################################### %

\subsection{Summary of Training Configurations}

Fig.~S1~in the Supplementary Materials summarizes the progression of the F1 score across all four training configurations on the ICU test set. \underline{Training Config \#1:} The zero-shot performance of both ECG foundation models is very high as is. \underline{Training Config \#2:} When the small ICU training set is used, the ECG foundation models and feature-based models perform better than the 1-D and 2-D deep learning models. \underline{Training Config \#3:} When the large non-ICU PhysioNet data is used, many of the deep learning and ECG foundation models perform better than the feature-based models. Therefore, there is a clear advantage to using the larger non-ICU PhysioNet data for training compared to the smaller ICU training set for the deep learning models trained from scratch. \underline{Training Config \#4:} Transfer learning improved performance in limited cases, such as ECGFounder.

% ##################################################### %

\subsection{Summary of AI Approaches}

\begin{figure*}
\centering
\includegraphics[width=\textwidth]{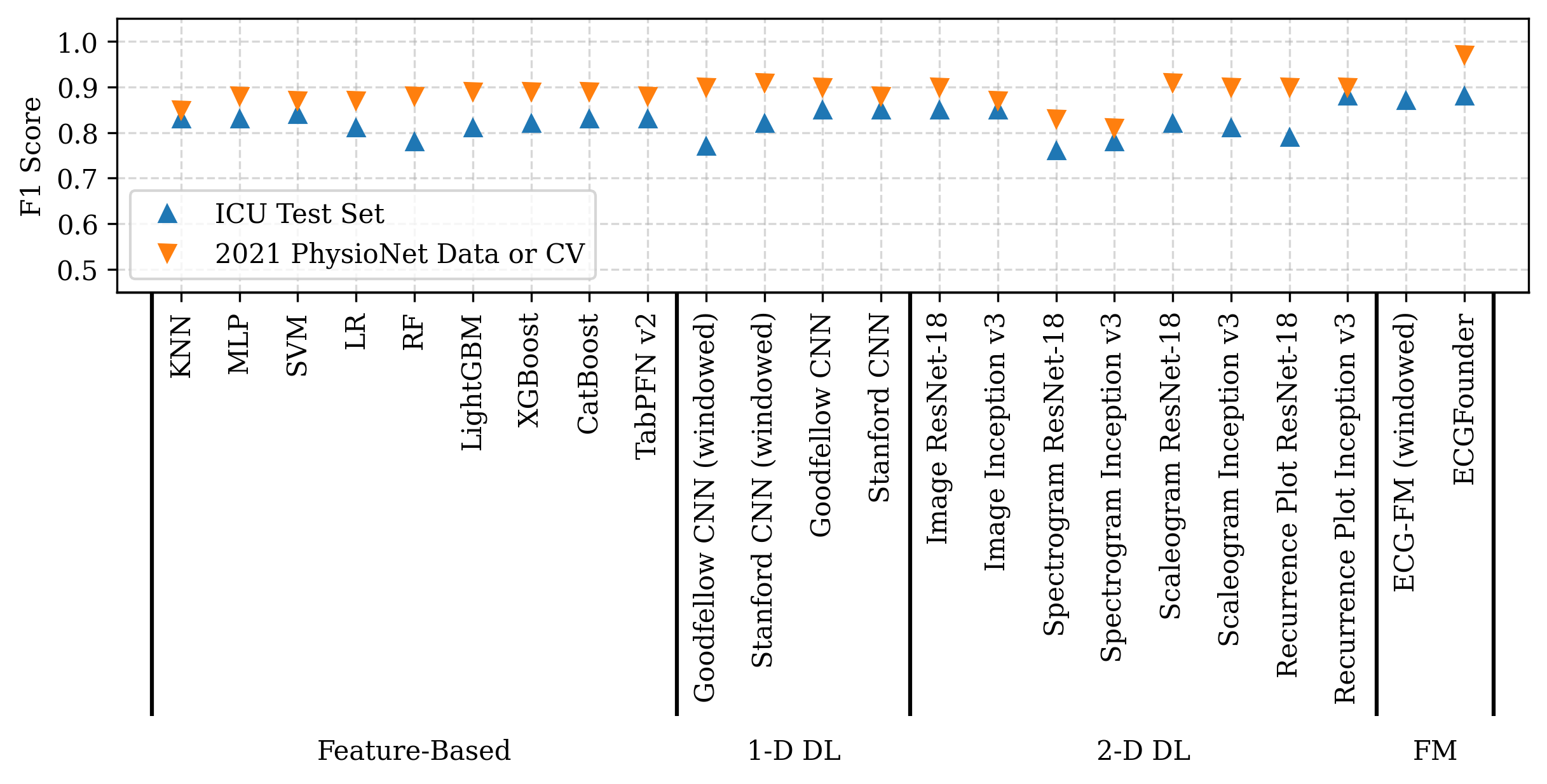}
\caption{Overall summary of best F1 scores across both datasets. Only the best F1 score across all training configurations is reported. ECG-FM is not included for the PhysioNet data since the pretrained model has already been exposed to this dataset.}
\label{figure:model_summary}
\end{figure*}

\begin{figure}
\centering
\includegraphics[width=\columnwidth]{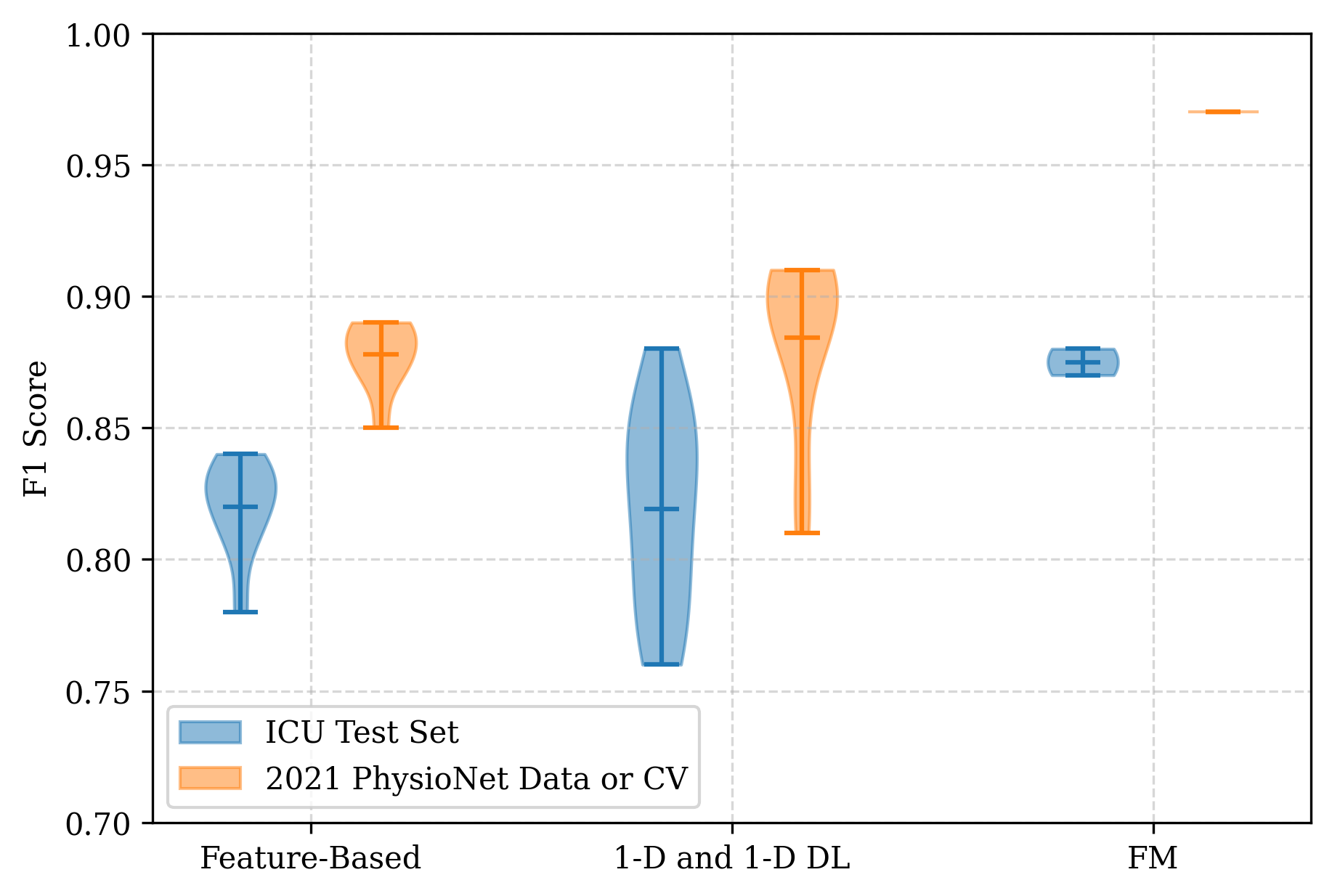}
\caption{Violin plot showing average, minimum, maximum, and distribution of the best F1 scores by AI approach across both datasets.}
\label{figure:approach_summary}
\end{figure}

Fig.~\ref{figure:model_summary}~shows the overall best F1 scores for each model on both the ICU test set and the 2021 PhysioNet Challenge data. The trends are similar for both datasets, but they most significantly differ with the random forest, windowed 1-D CNNs, scaleograms, and ResNet-18 with recurrence plots. At the higher performance levels, there is only a slight separation between top-performing models, indicating that the choice for which model to implement can rest on other factors, such as the precision metric to minimize alarm fatigue, inference speed, and computational requirements. Fig.~\ref{figure:approach_summary}~displays a violin plot showing the average, range, and distribution of the best F1 scores (across training configurations for each model) by general AI approach on both datasets. On average, feature-based models offer a very slight advantage over DL on our ICU test set, while DL offers more of an advantage on the 2021 PhysioNet dataset. Across both datasets, the F1 score of the DL methods more widely varies depending on the specific type of CNN model and input format, so while DL models sometimes outperformed the feature-based models, they also underperformed on certain occasions. It can also be observed that ECG FMs are in the lead and generally perform better than feature-based and DL models.

% ##################################################### %

\subsection{Confidence Intervals} \label{section:statistical_analysis}

Confidence intervals were calculated for the 23 models that achieved the best F1 scores reported in Fig.~\ref{figure:model_summary}. Each applicable test set (i.e., ICU test set, PhysioNet data, and PhysioNet CV folds) was bootstrapped 10,000 times, and the F1 scores were calculated for each model. The 95\% confidence intervals for each model's bootstrapped F1 scores are shown in Fig.~S2~in the Supplementary Materials.

% ##################################################### %

\subsection{Comparison to Related Work and Human Labelling Performance}

With respect to the two papers,~\cite{liaqat2020detection}~and~\cite{andreotti2017comparing}, comparing AI methods for AF detection that were discussed in the Related Work, our findings are similar but with a few key differences. First, our analysis included a wider range of methods in the comparison, including signal-to-image transformation models and ECG FMs. Second, one of the papers~\cite{liaqat2020detection}~found a much larger gap in performance between the best feature-based model and the best DL model (F1=0.700 vs F1=0.869). In contrast, this difference was not as pronounced in our experiments (F1=0.84 vs F1=0.88 on the ICU test set and F1=0.89 vs F1=0.91 with the 2021 PhysioNet Challenge data). This may come down to a difference in the set of features extracted, which would indicate that the performance of feature-based models is highly dependent on the quality of the hand-crafted features. Finally, the second paper~\cite{andreotti2017comparing}~found opposing trends depending on the evaluation set, where the best feature-based approach outperformed the best CNN approach with CV on the training set while the best CNN approach outperformed the best feature-based approach on the test set. In our comparison results, the trend that the best trained-from-scratch DL approach outperformed the best feature-based approach was consistent across both datasets. This difference may come down to the variety and amount of approaches considered, indicating that a larger-scale comparison experiment may lead to more consistent general trends across datasets.

Previous work using the Kingston ICU dataset reported an average F1 score of up to 0.78~\cite{chen2021detecting}~with 2.5~s windows and an F1 score of up to 0.67~\cite{chen2022quantifying}~or 0.714~\cite{chen2023deep}~with 10~s inputs using a 1-D CNN similar to the Goodfellow CNN used in this study. In our work, the maximum F1 score of the Goodfellow CNN with 2.5~s windows on our ICU test set was 0.77, and the maximum F1 score of the Goodfellow CNN with 10~s input was 0.85. As a result of performing a thorough comparison of different approaches, multiple models shown in Fig.~\ref{figure:model_summary}~outperformed what was reported in previous work, and several models that were directly compared with the Goodfellow CNN demonstrate improved performance on the ICU test set used here. The main differences between the preprocessing and training pipelines from this work and previous work are with regards to the signal filter, scaling, public dataset used, and Kingston ICU dataset split.

Several studies attempted to evaluate the performance of cardiologists and physicians in ECG interpretation. The Stanford CNN paper used 256-sample long inputs from single-lead ECG signals with a sampling frequency of 200~Hz, which corresponds to 1.28~s~\cite{hannun2019cardiologist}. In their paper, the authors used a dataset of 328 30~s ECG recordings from distinct patients as a test set. They reported an average cardiologist F1 score of 0.677 for the "Atrial Fibrillation \& Flutter" class when annotating every 1.28~s of input signal, compared to a model F1 score of 0.801. In an alternative testing configuration, the average cardiologist F1 score was 0.686 when identifying the rhythm's occurrence any time within the 30~s ECG recording, compared to a model F1 score of 0.831. What this analysis shows is that even cardiologists, who are expected to be the most skilled in ECG analysis, make frequent mistakes. AI models like the ones trained in our experiments therefore have the potential to not only make real-time rhythm classifications, but also make more correct classifications than cardiologists.

% ##################################################### %
% ##################################################### %
% ##################################################### %

\section{Limitations and Future Work}

This study focused on detecting AF versus sinus rhythm and not AF versus all other rhythms. For clinical deployment where an AI model would continuously monitor a patient's ECG and possibly encounter many rhythms as well as noise while making classifications in real time, it would be important to carefully consider how the model should be designed to handle recordings that present neither AF nor sinus rhythm instead of being forced to decide one class or the other. Techniques like model calibration, uncertainty estimation, or model training with data from all possible rhythm and noise scenarios could be used to allow the model to disclose when the rhythm does not fit into patterns learned from its training procedure. Future research can also combine the best models in an ensemble to further improve performance.

A limitation of the 2021 PhysioNet Challenge data is that ECG recordings sourced from certain sites are longer than 10~s and that many recordings have been assigned more than one label. Therefore, it is not clear whether multiple labels provided with a longer signal apply to the full length of the signal or whether the labels only indicate that a specific rhythm is present at any point in the signal. In addition, each source site from the 2021 PhysioNet Challenge data, along with the Kingston ICU dataset, had a unique annotation procedure and potentially differing paradigm for ECG interpretation. Further, the number of test samples in the Kingston ICU test set is relatively small. There are 298 test samples, only 50 of which were labelled as positive AF samples. A single misclassification in a small test set can have a larger impact on the evaluation metrics than a single misclassification in a large test set. All these factors signify the importance of unifying rhythm annotation definitions (e.g., whether atrial flutter should be grouped with AF), rigorous model validation across sites (e.g., leave-site-out cross-validation as performed in our research with the PhysioNet data), and benchmarking large datasets specifically for the task of AF detection in the ICU to permit direct comparisons among studies.

Although the feature-based approach underperformed the other two approaches in specific scenarios, this reflects the fundamental requirement of extracting and curating a set of features with strong predictive power, in contrast to DL and ECG FMs which can accept the raw data and automatically learn features during training. Thus, our feature-based models are limited by the amount, type, and quality of features we extracted. Our published ICU dataset along with our performance benchmarks enable the research community to continue advancing the state-of-the-art in this field by exploring other methods and models for AF detection and directly comparing performance on the same test set. One avenue of future work can be to improve the feature-based models, given their simplicity and interpretability, by extracting more features or improving the underlying peak detection algorithms.

There are two primary branches for future work to extend our research beyond improving AF detection performance on our published ICU dataset. The first branch is to prepare AF detectors suitable for clinical deployment, integrate them with ICU bedside monitors, and run clinical trials. Important considerations would be to monitor the model's performance compared to built-in bedside monitor algorithms and quantify the frequency of false alarms and prevalence of alarm fatigue in ICU staff with and without the AF detector. The second branch of future research is to explore AI approaches for the task of AF forecasting. For this study, we only used short 10~s randomly selected segments labelled by physicians. However, our data collection contains long-term recordings over patient stays with an average length of stay of 5.5 days. We intend to work with this larger collection for AF forecasting research and to utilize the models from this study as non-expert weak labellers for these long-term ECGs, since it would not be feasible for an expert to manually assess them to find AF onsets.

% ##################################################### %
% ##################################################### %
% ##################################################### %

\section{Conclusion}

The purpose of this research was to compare the performance of various AI techniques for AF detection from ECGs of ICU patients. These techniques ranged from simple feature-based ML to more complex DL models and recently published ECG FMs. The comparison experiments were conducted with an ICU dataset spanning around 600 patients, which we published, as well as public data from the 2021 PhysioNet Challenge. In general, ECG FMs outperformed feature-based and DL models in terms of F1 score, while the latter two approaches performed very similarly. The best F1 score on our ICU test set was 0.88 and was achieved by InceptionV3 with recurrence plots and a fine-tuned version of ECGFounder. By publishing an ICU dataset and thoroughly benchmarking AI approaches, this study aimed to advance the research of AI-powered AF detection in the ICU context and enable future research. Further, several high-performing detectors were trained and can be used as non-expert labellers to establish a foundation upon which research in AF forecasting can be conducted.

% ##################################################### %
% ##################################################### %
% ##################################################### %

\section{Acknowledgments}

This work was supported in part by the Natural Sciences and Engineering Research Council of Canada (NSERC), Social Sciences and Humanities Research Council (SSHRC), Kingston Health Sciences Centre (KHSC), and Vector Institute. S. Sibley, D. Pichora, and D. Maslove are members of the Attending Staff at KHSC. S. Sibley is supported in part by the Canadian Institutes of Health Research (CIHR), has received honoraria from Think Research, meeting sponsorships from Boston Scientific, Trimedic, and Icentia, and serves as a hospital organ donation physician with the Trillium Gift of Life Network at Ontario Health. D. Maslove is supported in part by the Southeastern Ontario Academic Medical Association (SEAMO). P. Mousavi is supported in part by a Canada CIFAR AI Chair and a Canada Research Chair.

% ##################################################### %
% ##################################################### %
% ##################################################### %

% References
\bibliographystyle{IEEEtran}
\bibliography{references}

% ##################################################### %
% ##################################################### %
% ##################################################### %

\clearpage
\onecolumn % Switches the layout to single column

% Reset counters and prefix them with "S"
\setcounter{equation}{0}
\setcounter{figure}{0}
\setcounter{table}{0}
\renewcommand{\thefigure}{S\arabic{figure}}
\renewcommand{\thetable}{S\arabic{table}}

\section*{Supplementary Matrials}

% -----------------------------------------------------

\begin{table}[ht]
\centering
\caption{ICU test set performance of zero-shot ECG FMs (training config \#1). Bold values indicate the best score per metric while underlined values indicate the second-best score per metric.}
\label{table:table_tc1_icutest_full}
\resizebox{\textwidth}{!}{
\begin{tabular}{|ccccccccccc|}
\hline
\multicolumn{1}{|c|}{Model} & \multicolumn{1}{c|}{AUROC ↑} & \multicolumn{1}{c|}{AUPRC ↑} & \multicolumn{1}{c|}{Log Loss ↓} & \multicolumn{1}{c|}{Brier ↓} & \multicolumn{1}{c|}{Recall ↑} & \multicolumn{1}{c|}{Precision ↑} & \multicolumn{1}{c|}{F1 ↑} & \multicolumn{1}{c|}{Specificity ↑} & \multicolumn{1}{c|}{\makecell{Balanced \\ Accuracy ↑}} & MCC ↑\\ \hline
\multicolumn{11}{|c|}{Zero-Shot FM} \\ \hline
\multicolumn{1}{|c|}{ECG-FM v1 (windowed)} & \multicolumn{1}{c|}{\textbf{0.98}} & \multicolumn{1}{c|}{\textbf{0.93}} & \multicolumn{1}{c|}{\textbf{0.14}} & \multicolumn{1}{c|}{\textbf{0.04}} & \multicolumn{1}{c|}{0.78} & \multicolumn{1}{c|}{\textbf{0.97}} & \multicolumn{1}{c|}{\textbf{0.86}} & \multicolumn{1}{c|}{\textbf{1.00}} & \multicolumn{1}{c|}{\underline{0.89}} & \textbf{0.85}\\
\multicolumn{1}{|c|}{ECG-FM v2 (windowed)} & \multicolumn{1}{c|}{0.96} & \multicolumn{1}{c|}{0.91} & \multicolumn{1}{c|}{0.45} & \multicolumn{1}{c|}{\underline{0.15}} & \multicolumn{1}{c|}{\textbf{0.92}} & \multicolumn{1}{c|}{0.42} & \multicolumn{1}{c|}{0.58} & \multicolumn{1}{c|}{0.76} & \multicolumn{1}{c|}{0.84} & 0.52\\
\multicolumn{1}{|c|}{12-Lead ECGFounder} & \multicolumn{1}{c|}{\underline{0.97}} & \multicolumn{1}{c|}{\underline{0.92}} & \multicolumn{1}{c|}{\underline{0.16}} & \multicolumn{1}{c|}{\textbf{0.04}} & \multicolumn{1}{c|}{0.78} & \multicolumn{1}{c|}{\underline{0.93}} & \multicolumn{1}{c|}{0.84} & \multicolumn{1}{c|}{\underline{0.99}} & \multicolumn{1}{c|}{0.88} & 0.82\\
\multicolumn{1}{|c|}{1-Lead ECGFounder} & \multicolumn{1}{c|}{\underline{0.97}} & \multicolumn{1}{c|}{\textbf{0.93}} & \multicolumn{1}{c|}{0.20} & \multicolumn{1}{c|}{\textbf{0.04}} & \multicolumn{1}{c|}{\underline{0.84}} & \multicolumn{1}{c|}{0.87} & \multicolumn{1}{c|}{\underline{0.85}} & \multicolumn{1}{c|}{0.98} & \multicolumn{1}{c|}{\textbf{0.91}} & \underline{0.83}\\ \hline
\end{tabular}}
\end{table}

% -----------------------------------------------------

\begin{table}[ht]
\centering
\caption{ICU test set performance of models trained on the ICU training set (training config \#2).}
\label{table:table_tc2_icutest_full}
\resizebox{\textwidth}{!}{
\begin{threeparttable}
\begin{tabular}{|ccccccccccc|}
\hline
\multicolumn{1}{|c|}{Model} & \multicolumn{1}{c|}{AUROC ↑} & \multicolumn{1}{c|}{AUPRC ↑} & \multicolumn{1}{c|}{Log Loss ↓} & \multicolumn{1}{c|}{Brier ↓} & \multicolumn{1}{c|}{Recall ↑} & \multicolumn{1}{c|}{Precision ↑} & \multicolumn{1}{c|}{F1 ↑} & \multicolumn{1}{c|}{Specificity ↑} & \multicolumn{1}{c|}{\makecell{Balanced \\ Accuracy ↑}} & MCC ↑\\ \hline
\multicolumn{11}{|c|}{Feature-Based} \\ \hline
\multicolumn{1}{|c|}{KNN} & \multicolumn{1}{c|}{0.95} & \multicolumn{1}{c|}{0.82} & \multicolumn{1}{c|}{0.70} & \multicolumn{1}{c|}{\underline{0.05}} & \multicolumn{1}{c|}{0.88} & \multicolumn{1}{c|}{0.78} & \multicolumn{1}{c|}{\underline{0.83}} & \multicolumn{1}{c|}{0.95} & \multicolumn{1}{c|}{\textbf{0.91}} & 0.79\\
\multicolumn{1}{|c|}{MLP} & \multicolumn{1}{c|}{0.96} & \multicolumn{1}{c|}{\underline{0.91}} & \multicolumn{1}{c|}{0.18} & \multicolumn{1}{c|}{\textbf{0.04}} & \multicolumn{1}{c|}{0.80} & \multicolumn{1}{c|}{0.87} & \multicolumn{1}{c|}{\underline{0.83}} & \multicolumn{1}{c|}{\underline{0.98}} & \multicolumn{1}{c|}{0.89} & \underline{0.80}\\
\multicolumn{1}{|c|}{SVM} & \multicolumn{1}{c|}{\underline{0.97}} & \multicolumn{1}{c|}{0.89} & \multicolumn{1}{c|}{\underline{0.16}} & \multicolumn{1}{c|}{\textbf{0.04}} & \multicolumn{1}{c|}{0.78} & \multicolumn{1}{c|}{0.84} & \multicolumn{1}{c|}{0.81} & \multicolumn{1}{c|}{0.97} & \multicolumn{1}{c|}{0.87} & 0.77\\
\multicolumn{1}{|c|}{LR} & \multicolumn{1}{c|}{\underline{0.97}} & \multicolumn{1}{c|}{0.88} & \multicolumn{1}{c|}{\underline{0.16}} & \multicolumn{1}{c|}{\underline{0.05}} & \multicolumn{1}{c|}{0.78} & \multicolumn{1}{c|}{0.83} & \multicolumn{1}{c|}{0.80} & \multicolumn{1}{c|}{0.97} & \multicolumn{1}{c|}{0.87} & 0.76\\
\multicolumn{1}{|c|}{RF} & \multicolumn{1}{c|}{0.96} & \multicolumn{1}{c|}{0.89} & \multicolumn{1}{c|}{0.28} & \multicolumn{1}{c|}{\underline{0.05}} & \multicolumn{1}{c|}{0.73} & \multicolumn{1}{c|}{0.84} & \multicolumn{1}{c|}{0.78} & \multicolumn{1}{c|}{0.97} & \multicolumn{1}{c|}{0.85} & 0.75\\
\multicolumn{1}{|c|}{LightGBM} & \multicolumn{1}{c|}{\underline{0.97}} & \multicolumn{1}{c|}{\textbf{0.92}} & \multicolumn{1}{c|}{0.32} & \multicolumn{1}{c|}{\underline{0.05}} & \multicolumn{1}{c|}{0.78} & \multicolumn{1}{c|}{0.84} & \multicolumn{1}{c|}{0.81} & \multicolumn{1}{c|}{0.97} & \multicolumn{1}{c|}{0.87} & 0.77\\
\multicolumn{1}{|c|}{XGBoost} & \multicolumn{1}{c|}{\underline{0.97}} & \multicolumn{1}{c|}{0.90} & \multicolumn{1}{c|}{0.18} & \multicolumn{1}{c|}{\underline{0.05}} & \multicolumn{1}{c|}{0.80} & \multicolumn{1}{c|}{0.85} & \multicolumn{1}{c|}{0.82} & \multicolumn{1}{c|}{0.97} & \multicolumn{1}{c|}{0.88} & 0.79\\
\multicolumn{1}{|c|}{CatBoost} & \multicolumn{1}{c|}{\underline{0.97}} & \multicolumn{1}{c|}{\underline{0.91}} & \multicolumn{1}{c|}{\textbf{0.15}} & \multicolumn{1}{c|}{\textbf{0.04}} & \multicolumn{1}{c|}{0.84} & \multicolumn{1}{c|}{0.82} & \multicolumn{1}{c|}{\underline{0.83}} & \multicolumn{1}{c|}{0.96} & \multicolumn{1}{c|}{\underline{0.90}} & 0.79\\
\multicolumn{1}{|c|}{TabPFN v2} & \multicolumn{1}{c|}{\underline{0.97}} & \multicolumn{1}{c|}{\textbf{0.92}} & \multicolumn{1}{c|}{\textbf{0.15}} & \multicolumn{1}{c|}{\textbf{0.04}} & \multicolumn{1}{c|}{0.78} & \multicolumn{1}{c|}{\underline{0.88}} & \multicolumn{1}{c|}{\underline{0.83}} & \multicolumn{1}{c|}{\underline{0.98}} & \multicolumn{1}{c|}{0.88} & \underline{0.80}\\ \hline
\multicolumn{11}{|c|}{1-D DL} \\ \hline
\multicolumn{1}{|c|}{Goodfellow CNN (windowed)} & \multicolumn{1}{c|}{0.62} & \multicolumn{1}{c|}{0.22} & \multicolumn{1}{c|}{0.44} & \multicolumn{1}{c|}{0.14} & \multicolumn{1}{c|}{0.00} & \multicolumn{1}{c|}{0.00*} & \multicolumn{1}{c|}{0.00*} & \multicolumn{1}{c|}{\textbf{1.00}} & \multicolumn{1}{c|}{0.50} & 0.00*\\
\multicolumn{1}{|c|}{Stanford CNN (windowed)} & \multicolumn{1}{c|}{0.65} & \multicolumn{1}{c|}{0.27} & \multicolumn{1}{c|}{0.43} & \multicolumn{1}{c|}{0.13} & \multicolumn{1}{c|}{0.00} & \multicolumn{1}{c|}{0.00*} & \multicolumn{1}{c|}{0.00*} & \multicolumn{1}{c|}{\textbf{1.00}} & \multicolumn{1}{c|}{0.50} & 0.00*\\
\multicolumn{1}{|c|}{Goodfellow CNN} & \multicolumn{1}{c|}{0.60} & \multicolumn{1}{c|}{0.23} & \multicolumn{1}{c|}{0.44} & \multicolumn{1}{c|}{0.14} & \multicolumn{1}{c|}{0.00} & \multicolumn{1}{c|}{0.00*} & \multicolumn{1}{c|}{0.00*} & \multicolumn{1}{c|}{\textbf{1.00}} & \multicolumn{1}{c|}{0.50} & 0.00*\\
\multicolumn{1}{|c|}{Stanford CNN} & \multicolumn{1}{c|}{0.63} & \multicolumn{1}{c|}{0.24} & \multicolumn{1}{c|}{0.44} & \multicolumn{1}{c|}{0.14} & \multicolumn{1}{c|}{0.00} & \multicolumn{1}{c|}{0.00*} & \multicolumn{1}{c|}{0.00*} & \multicolumn{1}{c|}{\textbf{1.00}} & \multicolumn{1}{c|}{0.50} & 0.00*\\ \hline
\multicolumn{11}{|c|}{2-D DL} \\ \hline
\multicolumn{1}{|c|}{Image ResNet-18} & \multicolumn{1}{c|}{0.50} & \multicolumn{1}{c|}{0.16} & \multicolumn{1}{c|}{13.32} & \multicolumn{1}{c|}{0.84} & \multicolumn{1}{c|}{\textbf{1.00}} & \multicolumn{1}{c|}{0.16} & \multicolumn{1}{c|}{0.28} & \multicolumn{1}{c|}{0.00} & \multicolumn{1}{c|}{0.50} & 0.00*\\
\multicolumn{1}{|c|}{Image Inception v3} & \multicolumn{1}{c|}{0.88} & \multicolumn{1}{c|}{0.50} & \multicolumn{1}{c|}{0.46} & \multicolumn{1}{c|}{0.11} & \multicolumn{1}{c|}{0.59} & \multicolumn{1}{c|}{0.54} & \multicolumn{1}{c|}{0.56} & \multicolumn{1}{c|}{0.90} & \multicolumn{1}{c|}{0.75} & 0.47\\
\multicolumn{1}{|c|}{Spectrogram ResNet-18} & \multicolumn{1}{c|}{0.66} & \multicolumn{1}{c|}{0.29} & \multicolumn{1}{c|}{1.96} & \multicolumn{1}{c|}{0.17} & \multicolumn{1}{c|}{0.06} & \multicolumn{1}{c|}{0.25} & \multicolumn{1}{c|}{0.10} & \multicolumn{1}{c|}{0.96} & \multicolumn{1}{c|}{0.51} & 0.05\\
\multicolumn{1}{|c|}{Spectrogram Inception v3} & \multicolumn{1}{c|}{0.50} & \multicolumn{1}{c|}{0.16} & \multicolumn{1}{c|}{2.62} & \multicolumn{1}{c|}{0.16} & \multicolumn{1}{c|}{0.00} & \multicolumn{1}{c|}{0.00*} & \multicolumn{1}{c|}{0.00*} & \multicolumn{1}{c|}{\textbf{1.00}} & \multicolumn{1}{c|}{0.50} & 0.00*\\
\multicolumn{1}{|c|}{Scaleogram ResNet-18} & \multicolumn{1}{c|}{0.58} & \multicolumn{1}{c|}{0.24} & \multicolumn{1}{c|}{2.62} & \multicolumn{1}{c|}{0.16} & \multicolumn{1}{c|}{0.00} & \multicolumn{1}{c|}{0.00*} & \multicolumn{1}{c|}{0.00*} & \multicolumn{1}{c|}{\textbf{1.00}} & \multicolumn{1}{c|}{0.50} & 0.00*\\
\multicolumn{1}{|c|}{Scaleogram Inception v3} & \multicolumn{1}{c|}{0.64} & \multicolumn{1}{c|}{0.25} & \multicolumn{1}{c|}{2.45} & \multicolumn{1}{c|}{0.16} & \multicolumn{1}{c|}{0.00} & \multicolumn{1}{c|}{0.00*} & \multicolumn{1}{c|}{0.00*} & \multicolumn{1}{c|}{\textbf{1.00}} & \multicolumn{1}{c|}{0.50} & 0.00*\\
\multicolumn{1}{|c|}{Recurrence Plot ResNet-18} & \multicolumn{1}{c|}{0.92} & \multicolumn{1}{c|}{0.80} & \multicolumn{1}{c|}{0.80} & \multicolumn{1}{c|}{0.14} & \multicolumn{1}{c|}{\underline{0.90}} & \multicolumn{1}{c|}{0.51} & \multicolumn{1}{c|}{0.65} & \multicolumn{1}{c|}{0.83} & \multicolumn{1}{c|}{0.86} & 0.59\\
\multicolumn{1}{|c|}{Recurrence Plot Inception v3} & \multicolumn{1}{c|}{0.19} & \multicolumn{1}{c|}{0.10} & \multicolumn{1}{c|}{1.24} & \multicolumn{1}{c|}{0.20} & \multicolumn{1}{c|}{0.00} & \multicolumn{1}{c|}{0.00} & \multicolumn{1}{c|}{0.00} & \multicolumn{1}{c|}{0.93} & \multicolumn{1}{c|}{0.46} & -0.11\\ \hline
\multicolumn{11}{|c|}{Fine-Tuned FM} \\ \hline
\multicolumn{1}{|c|}{ECG-FM (windowed)} & \multicolumn{1}{c|}{0.95} & \multicolumn{1}{c|}{0.90} & \multicolumn{1}{c|}{0.44} & \multicolumn{1}{c|}{\underline{0.05}} & \multicolumn{1}{c|}{0.78} & \multicolumn{1}{c|}{\underline{0.88}} & \multicolumn{1}{c|}{\underline{0.83}} & \multicolumn{1}{c|}{\underline{0.98}} & \multicolumn{1}{c|}{0.88} & \underline{0.80}\\
\multicolumn{1}{|c|}{ECGFounder} & \multicolumn{1}{c|}{\textbf{0.98}} & \multicolumn{1}{c|}{\textbf{0.92}} & \multicolumn{1}{c|}{0.45} & \multicolumn{1}{c|}{\textbf{0.04}} & \multicolumn{1}{c|}{0.82} & \multicolumn{1}{c|}{\textbf{0.89}} & \multicolumn{1}{c|}{\textbf{0.85}} & \multicolumn{1}{c|}{\underline{0.98}} & \multicolumn{1}{c|}{\underline{0.90}} & \textbf{0.82}\\ \hline
\end{tabular}
\begin{tablenotes}
\small
\item[*] Precision, F1 score, and MCC are mathematically undefined due to a zero denominator, but are conventionally bound to 0.00.
\end{tablenotes}
\end{threeparttable}}
\end{table}

% -----------------------------------------------------

\begin{table}[ht]
\centering
\caption{ICU test set performance of models trained on the 2021 PhysioNet Challenge data (training config \#3).}
\label{table:table_tc3_icutest_full}
\resizebox{\textwidth}{!}{
\begin{tabular}{|ccccccccccc|}
\hline
\multicolumn{1}{|c|}{Model} & \multicolumn{1}{c|}{AUROC ↑} & \multicolumn{1}{c|}{AUPRC ↑} & \multicolumn{1}{c|}{Log Loss ↓} & \multicolumn{1}{c|}{Brier ↓} & \multicolumn{1}{c|}{Recall ↑} & \multicolumn{1}{c|}{Precision ↑} & \multicolumn{1}{c|}{F1 ↑} & \multicolumn{1}{c|}{Specificity ↑} & \multicolumn{1}{c|}{\makecell{Balanced \\ Accuracy ↑}} & MCC ↑\\ \hline
\multicolumn{11}{|c|}{Feature-Based} \\ \hline
\multicolumn{1}{|c|}{KNN} & \multicolumn{1}{c|}{0.90} & \multicolumn{1}{c|}{0.74} & \multicolumn{1}{c|}{1.51} & \multicolumn{1}{c|}{0.06} & \multicolumn{1}{c|}{\underline{0.82}} & \multicolumn{1}{c|}{0.80} & \multicolumn{1}{c|}{0.81} & \multicolumn{1}{c|}{0.96} & \multicolumn{1}{c|}{0.89} & 0.77\\
\multicolumn{1}{|c|}{MLP} & \multicolumn{1}{c|}{0.93} & \multicolumn{1}{c|}{0.82} & \multicolumn{1}{c|}{0.24} & \multicolumn{1}{c|}{0.05} & \multicolumn{1}{c|}{0.73} & \multicolumn{1}{c|}{0.86} & \multicolumn{1}{c|}{0.79} & \multicolumn{1}{c|}{0.98} & \multicolumn{1}{c|}{0.86} & 0.76\\
\multicolumn{1}{|c|}{SVM} & \multicolumn{1}{c|}{0.93} & \multicolumn{1}{c|}{0.87} & \multicolumn{1}{c|}{0.17} & \multicolumn{1}{c|}{\underline{0.04}} & \multicolumn{1}{c|}{0.80} & \multicolumn{1}{c|}{0.89} & \multicolumn{1}{c|}{0.84} & \multicolumn{1}{c|}{0.98} & \multicolumn{1}{c|}{0.89} & 0.81\\
\multicolumn{1}{|c|}{LR} & \multicolumn{1}{c|}{\underline{0.97}} & \multicolumn{1}{c|}{0.88} & \multicolumn{1}{c|}{0.16} & \multicolumn{1}{c|}{0.05} & \multicolumn{1}{c|}{0.78} & \multicolumn{1}{c|}{0.84} & \multicolumn{1}{c|}{0.81} & \multicolumn{1}{c|}{0.97} & \multicolumn{1}{c|}{0.87} & 0.77\\
\multicolumn{1}{|c|}{RF} & \multicolumn{1}{c|}{0.93} & \multicolumn{1}{c|}{0.86} & \multicolumn{1}{c|}{0.21} & \multicolumn{1}{c|}{0.05} & \multicolumn{1}{c|}{0.71} & \multicolumn{1}{c|}{0.81} & \multicolumn{1}{c|}{0.76} & \multicolumn{1}{c|}{0.97} & \multicolumn{1}{c|}{0.84} & 0.72\\
\multicolumn{1}{|c|}{LightGBM} & \multicolumn{1}{c|}{0.93} & \multicolumn{1}{c|}{0.86} & \multicolumn{1}{c|}{0.33} & \multicolumn{1}{c|}{0.06} & \multicolumn{1}{c|}{0.76} & \multicolumn{1}{c|}{0.80} & \multicolumn{1}{c|}{0.78} & \multicolumn{1}{c|}{0.96} & \multicolumn{1}{c|}{0.86} & 0.74\\
\multicolumn{1}{|c|}{XGBoost} & \multicolumn{1}{c|}{0.94} & \multicolumn{1}{c|}{0.86} & \multicolumn{1}{c|}{0.39} & \multicolumn{1}{c|}{0.06} & \multicolumn{1}{c|}{0.78} & \multicolumn{1}{c|}{0.79} & \multicolumn{1}{c|}{0.78} & \multicolumn{1}{c|}{0.96} & \multicolumn{1}{c|}{0.87} & 0.74\\
\multicolumn{1}{|c|}{CatBoost} & \multicolumn{1}{c|}{0.94} & \multicolumn{1}{c|}{0.87} & \multicolumn{1}{c|}{0.21} & \multicolumn{1}{c|}{0.06} & \multicolumn{1}{c|}{0.78} & \multicolumn{1}{c|}{0.76} & \multicolumn{1}{c|}{0.77} & \multicolumn{1}{c|}{0.95} & \multicolumn{1}{c|}{0.86} & 0.72\\
\multicolumn{1}{|c|}{TabPFN v2} & \multicolumn{1}{c|}{0.94} & \multicolumn{1}{c|}{0.86} & \multicolumn{1}{c|}{0.21} & \multicolumn{1}{c|}{0.06} & \multicolumn{1}{c|}{0.73} & \multicolumn{1}{c|}{0.90} & \multicolumn{1}{c|}{0.81} & \multicolumn{1}{c|}{0.98} & \multicolumn{1}{c|}{0.86} & 0.78\\ \hline
\multicolumn{11}{|c|}{1-D DL} \\ \hline
\multicolumn{1}{|c|}{Goodfellow CNN (windowed)} & \multicolumn{1}{c|}{0.96} & \multicolumn{1}{c|}{0.79} & \multicolumn{1}{c|}{0.20} & \multicolumn{1}{c|}{0.06} & \multicolumn{1}{c|}{\underline{0.82}} & \multicolumn{1}{c|}{0.73} & \multicolumn{1}{c|}{0.77} & \multicolumn{1}{c|}{0.94} & \multicolumn{1}{c|}{0.88} & 0.72\\
\multicolumn{1}{|c|}{Stanford CNN (windowed)} & \multicolumn{1}{c|}{\textbf{0.98}} & \multicolumn{1}{c|}{0.91} & \multicolumn{1}{c|}{0.15} & \multicolumn{1}{c|}{0.05} & \multicolumn{1}{c|}{\underline{0.82}} & \multicolumn{1}{c|}{0.82} & \multicolumn{1}{c|}{0.82} & \multicolumn{1}{c|}{0.96} & \multicolumn{1}{c|}{0.89} & 0.78\\
\multicolumn{1}{|c|}{Goodfellow CNN} & \multicolumn{1}{c|}{\underline{0.97}} & \multicolumn{1}{c|}{0.92} & \multicolumn{1}{c|}{0.15} & \multicolumn{1}{c|}{\underline{0.04}} & \multicolumn{1}{c|}{\underline{0.82}} & \multicolumn{1}{c|}{0.89} & \multicolumn{1}{c|}{0.85} & \multicolumn{1}{c|}{0.98} & \multicolumn{1}{c|}{\underline{0.90}} & 0.82\\
\multicolumn{1}{|c|}{Stanford CNN} & \multicolumn{1}{c|}{\textbf{0.98}} & \multicolumn{1}{c|}{\underline{0.93}} & \multicolumn{1}{c|}{0.15} & \multicolumn{1}{c|}{\underline{0.04}} & \multicolumn{1}{c|}{0.80} & \multicolumn{1}{c|}{0.91} & \multicolumn{1}{c|}{0.85} & \multicolumn{1}{c|}{0.98} & \multicolumn{1}{c|}{0.89} & 0.82\\ \hline
\multicolumn{11}{|c|}{2-D DL} \\ \hline
\multicolumn{1}{|c|}{Image ResNet-18} & \multicolumn{1}{c|}{\textbf{0.98}} & \multicolumn{1}{c|}{\underline{0.93}} & \multicolumn{1}{c|}{0.14} & \multicolumn{1}{c|}{\underline{0.04}} & \multicolumn{1}{c|}{\textbf{0.84}} & \multicolumn{1}{c|}{0.87} & \multicolumn{1}{c|}{0.85} & \multicolumn{1}{c|}{0.98} & \multicolumn{1}{c|}{\textbf{0.91}} & 0.83\\
\multicolumn{1}{|c|}{Image Inception v3} & \multicolumn{1}{c|}{\textbf{0.98}} & \multicolumn{1}{c|}{\underline{0.93}} & \multicolumn{1}{c|}{\underline{0.12}} & \multicolumn{1}{c|}{\underline{0.04}} & \multicolumn{1}{c|}{0.80} & \multicolumn{1}{c|}{0.91} & \multicolumn{1}{c|}{0.85} & \multicolumn{1}{c|}{0.98} & \multicolumn{1}{c|}{0.89} & 0.82\\
\multicolumn{1}{|c|}{Spectrogram ResNet-18} & \multicolumn{1}{c|}{0.93} & \multicolumn{1}{c|}{0.77} & \multicolumn{1}{c|}{0.23} & \multicolumn{1}{c|}{0.07} & \multicolumn{1}{c|}{0.53} & \multicolumn{1}{c|}{0.87} & \multicolumn{1}{c|}{0.66} & \multicolumn{1}{c|}{0.98} & \multicolumn{1}{c|}{0.76} & 0.63\\
\multicolumn{1}{|c|}{Spectrogram Inception v3} & \multicolumn{1}{c|}{0.96} & \multicolumn{1}{c|}{0.77} & \multicolumn{1}{c|}{0.19} & \multicolumn{1}{c|}{0.06} & \multicolumn{1}{c|}{0.78} & \multicolumn{1}{c|}{0.78} & \multicolumn{1}{c|}{0.78} & \multicolumn{1}{c|}{0.96} & \multicolumn{1}{c|}{0.87} & 0.73\\
\multicolumn{1}{|c|}{Scaleogram ResNet-18} & \multicolumn{1}{c|}{\textbf{0.98}} & \multicolumn{1}{c|}{0.92} & \multicolumn{1}{c|}{0.19} & \multicolumn{1}{c|}{0.05} & \multicolumn{1}{c|}{0.65} & \multicolumn{1}{c|}{0.91} & \multicolumn{1}{c|}{0.76} & \multicolumn{1}{c|}{\underline{0.99}} & \multicolumn{1}{c|}{0.82} & 0.74\\
\multicolumn{1}{|c|}{Scaleogram Inception v3} & \multicolumn{1}{c|}{\underline{0.97}} & \multicolumn{1}{c|}{0.91} & \multicolumn{1}{c|}{0.16} & \multicolumn{1}{c|}{0.05} & \multicolumn{1}{c|}{0.78} & \multicolumn{1}{c|}{0.84} & \multicolumn{1}{c|}{0.81} & \multicolumn{1}{c|}{0.97} & \multicolumn{1}{c|}{0.87} & 0.77\\
\multicolumn{1}{|c|}{Recurrence Plot ResNet-18} & \multicolumn{1}{c|}{\textbf{0.98}} & \multicolumn{1}{c|}{0.92} & \multicolumn{1}{c|}{0.17} & \multicolumn{1}{c|}{0.05} & \multicolumn{1}{c|}{\textbf{0.84}} & \multicolumn{1}{c|}{0.75} & \multicolumn{1}{c|}{0.79} & \multicolumn{1}{c|}{0.94} & \multicolumn{1}{c|}{0.89} & 0.75\\
\multicolumn{1}{|c|}{Recurrence Plot Inception v3} & \multicolumn{1}{c|}{\underline{0.97}} & \multicolumn{1}{c|}{0.92} & \multicolumn{1}{c|}{0.13} & \multicolumn{1}{c|}{\underline{0.04}} & \multicolumn{1}{c|}{0.80} & \multicolumn{1}{c|}{\textbf{0.98}} & \multicolumn{1}{c|}{\textbf{0.88}} & \multicolumn{1}{c|}{\textbf{1.00}} & \multicolumn{1}{c|}{\underline{0.90}} & \textbf{0.86}\\ \hline
\multicolumn{11}{|c|}{Fine-Tuned FM} \\ \hline
\multicolumn{1}{|c|}{ECG-FM (windowed)} & \multicolumn{1}{c|}{\textbf{0.98}} & \multicolumn{1}{c|}{\textbf{0.94}} & \multicolumn{1}{c|}{\textbf{0.11}} & \multicolumn{1}{c|}{\textbf{0.03}} & \multicolumn{1}{c|}{\textbf{0.84}} & \multicolumn{1}{c|}{0.91} & \multicolumn{1}{c|}{\underline{0.87}} & \multicolumn{1}{c|}{0.98} & \multicolumn{1}{c|}{\textbf{0.91}} & \underline{0.85}\\
\multicolumn{1}{|c|}{ECGFounder} & \multicolumn{1}{c|}{\underline{0.97}} & \multicolumn{1}{c|}{0.92} & \multicolumn{1}{c|}{0.14} & \multicolumn{1}{c|}{\underline{0.04}} & \multicolumn{1}{c|}{0.78} & \multicolumn{1}{c|}{\underline{0.93}} & \multicolumn{1}{c|}{0.84} & \multicolumn{1}{c|}{\underline{0.99}} & \multicolumn{1}{c|}{0.88} & 0.82\\ \hline
\end{tabular}}
\end{table}

% -----------------------------------------------------

\begin{table}[ht]
\centering
\caption{ICU test set performance of models trained on the 2021 PhysioNet Challenge data then fine-tuned with the ICU training set for transfer learning (training config \#4).}
\label{table:table_tc4_icutest_full}
\resizebox{\textwidth}{!}{
\begin{tabular}{|ccccccccccc|}
\hline
\multicolumn{1}{|c|}{Model} & \multicolumn{1}{c|}{AUROC ↑} & \multicolumn{1}{c|}{AUPRC ↑} & \multicolumn{1}{c|}{Log Loss ↓} & \multicolumn{1}{c|}{Brier ↓} & \multicolumn{1}{c|}{Recall ↑} & \multicolumn{1}{c|}{Precision ↑} & \multicolumn{1}{c|}{F1 ↑} & \multicolumn{1}{c|}{Specificity ↑} & \multicolumn{1}{c|}{\makecell{Balanced \\ Accuracy ↑}} & MCC ↑\\ \hline
\multicolumn{11}{|c|}{1-D DL} \\ \hline
\multicolumn{1}{|c|}{Goodfellow CNN (windowed)} & \multicolumn{1}{c|}{0.94} & \multicolumn{1}{c|}{0.77} & \multicolumn{1}{c|}{0.27} & \multicolumn{1}{c|}{0.07} & \multicolumn{1}{c|}{0.53} & \multicolumn{1}{c|}{0.81} & \multicolumn{1}{c|}{0.64} & \multicolumn{1}{c|}{0.98} & \multicolumn{1}{c|}{0.75} & 0.61\\
\multicolumn{1}{|c|}{Stanford CNN (windowed)} & \multicolumn{1}{c|}{\underline{0.97}} & \multicolumn{1}{c|}{0.89} & \multicolumn{1}{c|}{\textbf{0.16}} & \multicolumn{1}{c|}{0.05} & \multicolumn{1}{c|}{0.73} & \multicolumn{1}{c|}{0.90} & \multicolumn{1}{c|}{0.81} & \multicolumn{1}{c|}{0.98} & \multicolumn{1}{c|}{0.86} & 0.78\\
\multicolumn{1}{|c|}{Goodfellow CNN} & \multicolumn{1}{c|}{\textbf{0.98}} & \multicolumn{1}{c|}{\textbf{0.94}} & \multicolumn{1}{c|}{\underline{0.18}} & \multicolumn{1}{c|}{\underline{0.04}} & \multicolumn{1}{c|}{0.84} & \multicolumn{1}{c|}{0.87} & \multicolumn{1}{c|}{0.85} & \multicolumn{1}{c|}{0.98} & \multicolumn{1}{c|}{\textbf{0.91}} & \underline{0.83}\\
\multicolumn{1}{|c|}{Stanford CNN} & \multicolumn{1}{c|}{\underline{0.97}} & \multicolumn{1}{c|}{0.90} & \multicolumn{1}{c|}{0.32} & \multicolumn{1}{c|}{0.06} & \multicolumn{1}{c|}{\textbf{0.88}} & \multicolumn{1}{c|}{0.77} & \multicolumn{1}{c|}{0.82} & \multicolumn{1}{c|}{0.95} & \multicolumn{1}{c|}{\textbf{0.91}} & 0.78\\ \hline
\multicolumn{11}{|c|}{2-D DL} \\ \hline
\multicolumn{1}{|c|}{Image ResNet-18} & \multicolumn{1}{c|}{\underline{0.97}} & \multicolumn{1}{c|}{\underline{0.91}} & \multicolumn{1}{c|}{0.48} & \multicolumn{1}{c|}{0.06} & \multicolumn{1}{c|}{0.78} & \multicolumn{1}{c|}{0.83} & \multicolumn{1}{c|}{0.80} & \multicolumn{1}{c|}{0.97} & \multicolumn{1}{c|}{0.87} & 0.76\\
\multicolumn{1}{|c|}{Image Inception v3} & \multicolumn{1}{c|}{\textbf{0.98}} & \multicolumn{1}{c|}{0.89} & \multicolumn{1}{c|}{0.54} & \multicolumn{1}{c|}{0.06} & \multicolumn{1}{c|}{0.69} & \multicolumn{1}{c|}{\underline{0.92}} & \multicolumn{1}{c|}{0.79} & \multicolumn{1}{c|}{\underline{0.99}} & \multicolumn{1}{c|}{0.84} & 0.77\\
\multicolumn{1}{|c|}{Spectrogram ResNet-18} & \multicolumn{1}{c|}{\underline{0.97}} & \multicolumn{1}{c|}{0.86} & \multicolumn{1}{c|}{0.46} & \multicolumn{1}{c|}{0.07} & \multicolumn{1}{c|}{0.69} & \multicolumn{1}{c|}{0.85} & \multicolumn{1}{c|}{0.76} & \multicolumn{1}{c|}{0.98} & \multicolumn{1}{c|}{0.83} & 0.73\\
\multicolumn{1}{|c|}{Spectrogram Inception v3} & \multicolumn{1}{c|}{0.94} & \multicolumn{1}{c|}{0.76} & \multicolumn{1}{c|}{0.66} & \multicolumn{1}{c|}{0.09} & \multicolumn{1}{c|}{0.65} & \multicolumn{1}{c|}{0.73} & \multicolumn{1}{c|}{0.69} & \multicolumn{1}{c|}{0.95} & \multicolumn{1}{c|}{0.80} & 0.63\\
\multicolumn{1}{|c|}{Scaleogram ResNet-18} & \multicolumn{1}{c|}{\textbf{0.98}} & \multicolumn{1}{c|}{0.89} & \multicolumn{1}{c|}{0.51} & \multicolumn{1}{c|}{0.05} & \multicolumn{1}{c|}{0.78} & \multicolumn{1}{c|}{0.86} & \multicolumn{1}{c|}{0.82} & \multicolumn{1}{c|}{0.98} & \multicolumn{1}{c|}{0.88} & 0.79\\
\multicolumn{1}{|c|}{Scaleogram Inception v3} & \multicolumn{1}{c|}{\underline{0.97}} & \multicolumn{1}{c|}{0.89} & \multicolumn{1}{c|}{0.37} & \multicolumn{1}{c|}{0.05} & \multicolumn{1}{c|}{0.80} & \multicolumn{1}{c|}{0.81} & \multicolumn{1}{c|}{0.80} & \multicolumn{1}{c|}{0.96} & \multicolumn{1}{c|}{0.88} & 0.77\\
\multicolumn{1}{|c|}{Recurrence Plot ResNet-18} & \multicolumn{1}{c|}{0.95} & \multicolumn{1}{c|}{0.87} & \multicolumn{1}{c|}{0.44} & \multicolumn{1}{c|}{0.06} & \multicolumn{1}{c|}{0.78} & \multicolumn{1}{c|}{0.81} & \multicolumn{1}{c|}{0.79} & \multicolumn{1}{c|}{0.96} & \multicolumn{1}{c|}{0.87} & 0.75\\
\multicolumn{1}{|c|}{Recurrence Plot Inception v3} & \multicolumn{1}{c|}{\underline{0.97}} & \multicolumn{1}{c|}{0.89} & \multicolumn{1}{c|}{0.42} & \multicolumn{1}{c|}{0.05} & \multicolumn{1}{c|}{0.84} & \multicolumn{1}{c|}{0.84} & \multicolumn{1}{c|}{0.84} & \multicolumn{1}{c|}{0.97} & \multicolumn{1}{c|}{\underline{0.90}} & 0.80\\ \hline
\multicolumn{11}{|c|}{Fine-Tuned FM} \\ \hline
\multicolumn{1}{|c|}{ECG-FM (windowed)} & \multicolumn{1}{c|}{\textbf{0.98}} & \multicolumn{1}{c|}{\textbf{0.94}} & \multicolumn{1}{c|}{0.25} & \multicolumn{1}{c|}{\underline{0.04}} & \multicolumn{1}{c|}{\underline{0.86}} & \multicolumn{1}{c|}{0.86} & \multicolumn{1}{c|}{\underline{0.86}} & \multicolumn{1}{c|}{0.97} & \multicolumn{1}{c|}{\textbf{0.91}} & \underline{0.83}\\
\multicolumn{1}{|c|}{ECGFounder} & \multicolumn{1}{c|}{\textbf{0.98}} & \multicolumn{1}{c|}{\textbf{0.94}} & \multicolumn{1}{c|}{0.35} & \multicolumn{1}{c|}{\textbf{0.03}} & \multicolumn{1}{c|}{0.80} & \multicolumn{1}{c|}{\textbf{0.98}} & \multicolumn{1}{c|}{\textbf{0.88}} & \multicolumn{1}{c|}{\textbf{1.00}} & \multicolumn{1}{c|}{\underline{0.90}} & \textbf{0.86}\\ \hline
\end{tabular}}
\end{table}

% -----------------------------------------------------

\begin{figure*}[ht]
\centering
\includegraphics[width=\textwidth]{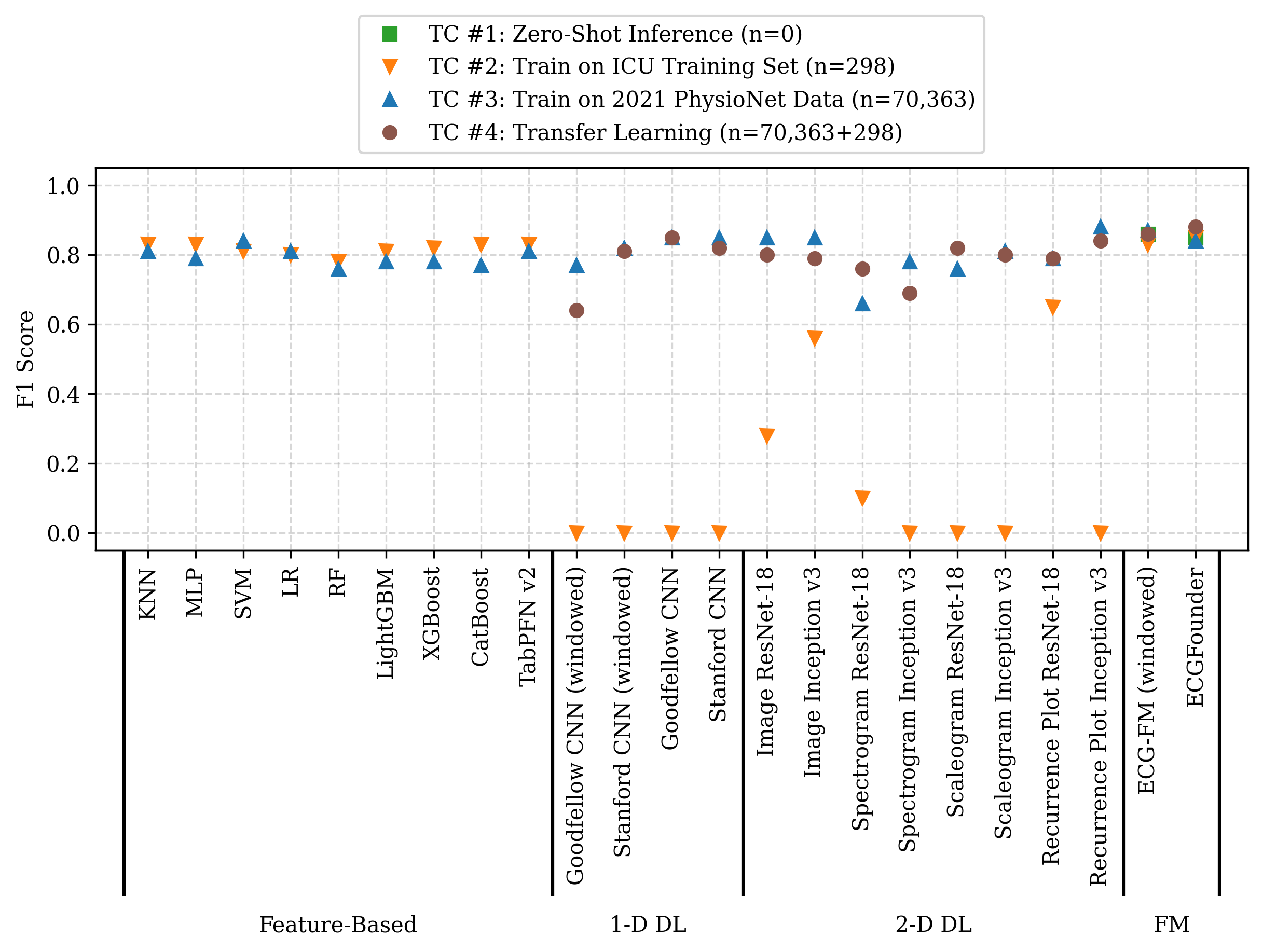}
\caption{Summary of F1 score on ICU test set across training configurations (TC: Training Configuration). The two entries for TC\#1 are present at the rightmost of the plot.}
\label{figure:tc_summary_icutest}
\end{figure*}

% -----------------------------------------------------

\begin{figure*}[ht]
\centering
\includegraphics[width=\textwidth]{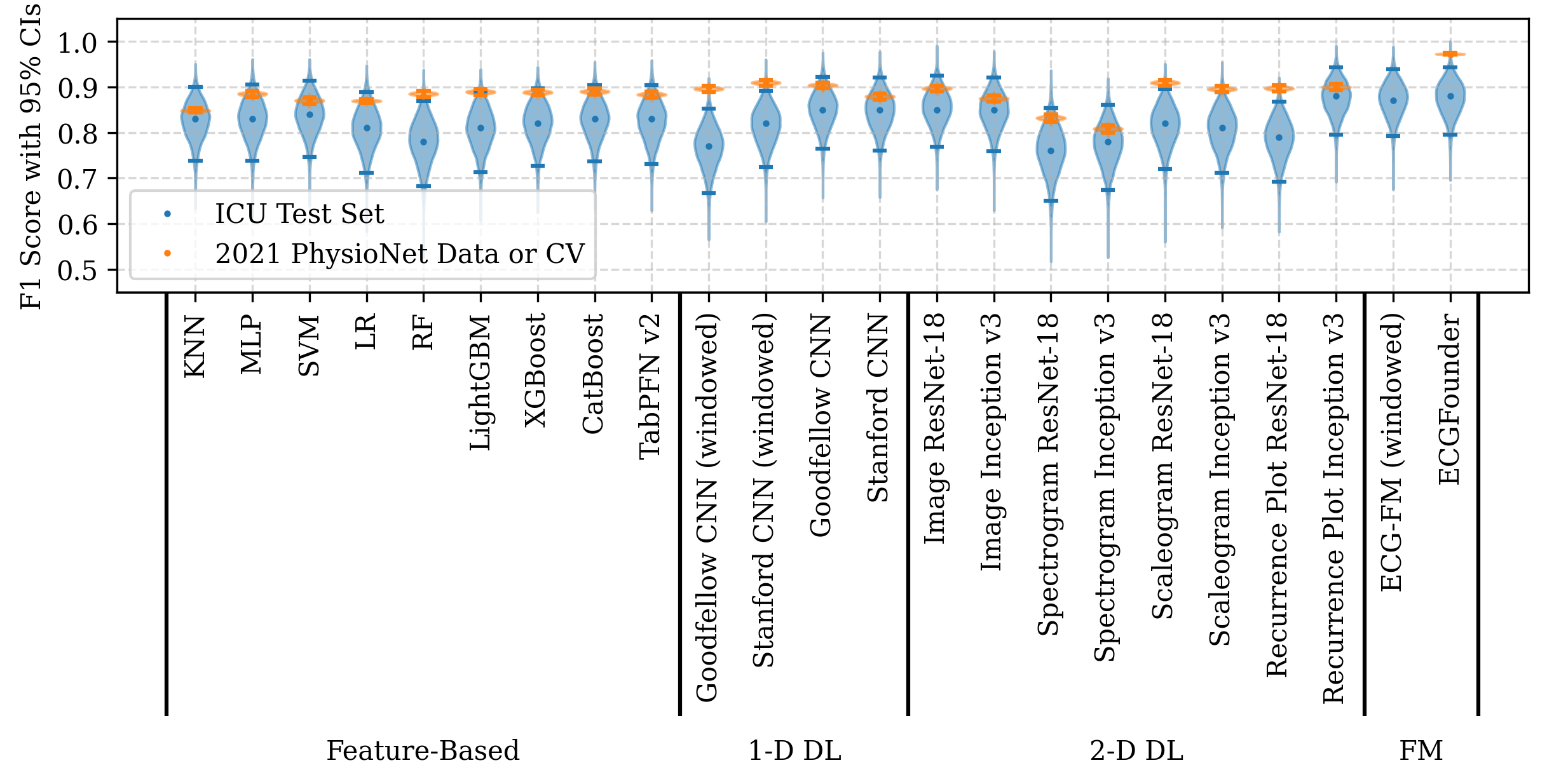}
\caption{Overall summary of best F1 scores across both datasets with confidence intervals (CIs). Only the best F1 score across all training configurations is reported. ECG-FM is not included for the PhysioNet data since the pretrained model has already been exposed to this dataset. CIs on the PhysioNet data are much tighter due to the larger sample size.}
\label{figure:model_summary_CIs}
\end{figure*}

% -----------------------------------------------------

% ##################################################### %
% ##################################################### %
% ##################################################### %

\end{document}